\documentclass[11pt]{article}

\usepackage[title]{appendix}

\usepackage{tikz}
\usepackage{amsmath}

\usepackage{amsfonts}
\usepackage{amsthm}
\usepackage{bigints}
\usepackage{booktabs}
\usepackage{caption} 
\usepackage{enumitem}
\usepackage{float}
\usepackage{mathtools}
\usepackage{subfigure}
\usepackage{hyperref}
\usepackage{wrapfig}
\newtheorem*{theorem*}{Theorem}

\usepackage{multirow}
\graphicspath{ {./images/} }
\usepackage{algorithm,algpseudocode}

\algnewcommand{\Inputs}[1]{%
  \State \textbf{Inputs:}
  \Statex \hspace*{\algorithmicindent}\parbox[t]{\linewidth}{\raggedright #1}
}
\algnewcommand{\Output}[1]{%
  \State \textbf{Output:}
  \Statex \hspace*{\algorithmicindent}\parbox[t]{\linewidth}{\raggedright #1}
}
\algnewcommand{\Initialize}[1]{%
  \State \textbf{Initialize:}
  \Statex \hspace*{\algorithmicindent}\parbox[t]{\linewidth}{\raggedright #1}
}
\algnewcommand{\comm}[1]{ {\ttfamily\textcolor{blue}{// #1}} }
\algnewcommand{\commblack}[1]{ {\ttfamily{// #1}} }
\PassOptionsToPackage{hyphens}{url}\usepackage{hyperref}
\DeclareMathOperator*{\argmax}{arg\,max}

\usepackage{titletoc}

\usepackage{filecontents}

\begin{document}

\date{}

\title{\Large \bf Correlation inference attacks against machine learning models}

\renewcommand{\thefootnote}{\fnsymbol{footnote}}

\author{
{\rm Ana-Maria Cre\c{t}u$^\dagger$}
\\
EPFL \\
ana-maria.cretu@epfl.ch
\and
{\rm Florent Gu\'epin$^\dagger$}
\\
Imperial College London \\
florent.guepin@imperial.ac.uk
\and
 {\rm Yves-Alexandre de Montjoye$^*$}\\
Imperial College London \\
demontjoye@imperial.ac.uk 
\\\\
$^\dagger$ These authors contributed equally to this work.
} 

\maketitle


\begin{abstract}

Despite machine learning models being widely used today, the relationship between a model and its training dataset is not well understood. 
We explore correlation inference attacks, whether and when a model leaks information about the correlations between the input variables of its training dataset. We first propose a \textit{model-less} attack, where an adversary exploits the spherical parametrization of correlation matrices alone to make an informed guess. Second, we propose a \textit{model-based} attack, where an adversary exploits black-box model access to infer the correlations using minimal and realistic assumptions. Third, we evaluate our attacks against logistic regression and multilayer perceptron models on three tabular datasets and show the models to leak correlations.
We finally show how extracted correlations can be used as building blocks for attribute inference attacks and enable weaker adversaries.
Our results raise fundamental questions on what a model does and should remember from its training set.

\end{abstract}

\section{Introduction}
Machine learning (ML) is widely used today by companies, governments, and researchers to learn from data how to automate processes and make decisions. Machine learning models are, for example, used for medical diagnosis~\cite{qummar2019deep}, language translation~\cite{wu2016google}, speech recognition and synthesis~\cite{siri2017}, chatbots, content moderation~\cite{amazonrekognition2018}, drug discovery~\cite{chen2018rise}, content retrieval~\cite{gordo2016deep}, and personalized services.

While models are often described as solutions to general tasks, e.g., detecting early-stage cancer in young patients, learning is performed on specific finite datasets~\cite{goodfellow2016deep}. The understanding of the relationship between a model and the specific dataset it was trained on is both a fundamental question, from a learning theory perspective, and a very important one in practice, for instance to evaluate the circumstances in which a model can be used. The extreme cases would be on the one hand an ideal hypothetical case where the model remembers nothing more from the dataset what is necessary for its intended goal: inferring an output variable based on the input – what we write $\mathcal{M}(x) \simeq P(Y|X=x)$. The other extreme would be a case where the model encodes the entire dataset and, when asked for a prediction, outputs the label of the closest record in its training set. Both are obviously extreme hypothetical cases and real models are likely to fall somewhere in between. For instance, models have until recently been smaller in size than their training dataset, suggesting that some level of learning happens. On the other hand, million- and even billion-parameter models have today become standard, e.g., in large language models~\cite{microsoft2021}, suggesting that models might remember quite a lot about their training dataset potentially bringing us slightly closer to the encoded case. 

The intuitive way to study the relationship between a model and its training dataset could be to track how the model ``came to be'', i.e., how the final parameters of the model are computed, step by step, from the training dataset. This is, however, increasingly difficult. Learning from a dataset today indeed involves performing a complex, iterative, and stochastic procedure, over random subsets of the dataset (so-called mini-batches) to update sometimes billions of parameters using gradient descent. In practice, the resulting model is closer to a ``black box'': a set of parameters whose precise relationship to the training dataset is largely untraceable.

Trained models have instead been studied in a post-hoc fashion, trying to understand what they can reveal about their training dataset, and thus what they ``remember''~\cite{veale2018algorithms}. More specifically, inference attacks, in the form of games, have been used to study whether specific pieces of information can be extracted from a trained model, showing that the model ``remembers'' them~\cite{ateniese2015hacking,fredrikson2014privacy,fredrikson2015model,shokri2017membership,carlini2021extracting,balle2022reconstructing}. 

So far, inference attacks have however mostly focused on inferring the presence or absence of a given record in the dataset, a membership inference attack (MIA)~\cite{homer2008resolving,shokri2017membership}. In an MIA, the \emph{adversary} aims to infer, better than random, whether a specific record was part of a dataset used to train the model. A large body of work has used MIAs to build a robust understanding of the conditions under which information about individual records is more likely to leak, e.g. when a model is overfitted or trained on a small dataset~\cite{salem2018ml,yeom2018privacy,nasr2019comprehensive,truex2019demystifying,jayaraman2019evaluating,leino2020stolen,choquette2021label,carlini2022membership}.

Membership is, however, only one of the characteristics of the training dataset that can be remembered by a model. We here exclude the hypothetical case where membership attacks would achieve perfect or close to perfect accuracy and records can be enumerated. This would lead to the dataset being reconstructable and any other properties to be directly computed. Current MIAs are however very far from reaching a perfect accuracy across a large range of records. We do not believe that near-future attacks are likely to reach close to perfect accuracy in general either.

In particular, there has been little work studying whether models encode dataset-level characteristics. An important line of research, for instance, aims to audit machine learning models to study biases or to better understand the range of values within which a model can operate, a crucial question in practice~\cite{buolamwini2018gender}. From a privacy perspective, dataset-level characteristics could reveal sensitive information~\cite{ateniese2015hacking,ganju2018property,zhang2021leakage}. For instance, a scoring model for depression could remember that patients living in the inner city are more likely to have used illegal substances.
The answer to the question of what a model does or even should remember from its training dataset is not simple. One could argue that the ideal ``fully generalizable'' model should learn nothing more than the intended probability of the output label given an input sample $\mathcal{M}(x) \simeq P(Y|X=x)$. Models today and in the near future, however, result from minimizing a loss function computed over a specific finite dataset of samples from the underlying dataset distribution $P(X,Y)$~\cite{goodfellow2016deep}. The dataset distribution $P(X, Y)$ and, through marginalization, $P(X)$, may very well end up influencing the final model in practice. 

\textbf{Contribution.} In this work, we study a new type of leakage in ML models: the leakage of correlations between input variables of a tabular training dataset by proposing what is, to our knowledge, the first correlation inference attack (CIA) against ML models.
We argue that (a) the leakage of correlations by an ML model can be seen as a privacy violation which goes beyond the model's intended purpose and can lead to individual harm (see Sec.~\ref{sec:discussion} for examples), (b) correlations can be used as building blocks for other attacks on individual privacy such as attribute inference attacks, in particular enabling weaker adversaries (see Sec.~\ref{subsec:results:attribute-inference-attack}), and (c) CIAs are different from previously studied property inference attacks (PIA), since PIAs target something else and rely on strong assumptions that make them inapplicable to CIAs (see Sec.~\ref{sec:related-work} for a detailed comparison).

We propose two different correlation inference attacks. First, a model-less attack that uses knowledge of the correlations between the input variables $X$ and the output variable $Y$, which we call correlation constraints, to predict the correlation between two input variables $\rho(X_1, X_2)$. We argue that the correlation constraints are typically known to or inferrable by the attacker. Our attack exploits the spherical parametrization of correlation matrices, imposing bounds on the correlation coefficients, to make an informed guess. This first attack also acts as a strong baseline, allowing us to correctly quantify the leakage caused by the machine learning model, and not by the adversary’s prior knowledge about the dataset. We then propose a second, model-based attack which exploits access to an ML model trained on the dataset of interest.
Our model-based attack uses a meta-classifier relying only on synthetic data, dramatically reducing the strength of the adversary compared to the literature. To generate synthetic data, we combine Gaussian copula-based generative modeling as our prior with carefully adapted procedures for sampling correlation matrices under constraints.  

We then study the performance of our model-based attack on synthetic data under three different attack scenarios. By order of increasing adversary knowledge, we consider an adversary that knows: (1) the correlation constraints relating only to the target variables, (2) all the input variables, or (3) the correlations between all the variables except for the target correlation, $\rho(X_1, X_2)$. Across scenarios, we first show our model-based attack to consistently strongly outperform the model-less attack, showing that ML models such as Logistic Regressions (LR) and Multilayer Perceptrons (MLP) leak information about the correlations between their input variables $X$. Second, we show how mitigations such as limiting the number of queries that can be made to the model or their precision fail to prevent the attack. Third, we show that, unsurprisingly, ensuring that the model is differentially private (DP) does not prevent our attack and extend our attack to and validate it  on three real-world datasets. 
We finally argue that dataset-level statistics can be building blocks for other attacks, in particular enabling weaker adversaries. To illustrate our argument, we propose a novel attribute inference attack exploiting the correlations extracted from the model, and show it to significantly outperform previous attacks on real-world datasets. 

We make our source code allowing to reproduce the results of our experiments available at: \url{https://github.com/computationalprivacy/ml-correlation-inference}.

\section{Results}
\subsection{Correlation inference threat model}
\label{subsec:results:attack-model}
We consider a machine learning model $\mathcal{M}_T$ trained on a \textit{target dataset} $D_{T}$ consisting of $m$ records $z^1, \ldots, z^m$. Each $z^i$ is a sample of $n-1$ input variables $X_1, \ldots, X_{n-1}$ and one output variable $Y$. We assume variables to be continuous and real-valued and denote by $F_i(x) = \text{Pr}(X_i \leq x)$ the one-way marginal of the $i$-th input variable and by $F_n(y)$ the one-way marginal of the output variable.
For simplicity, we consider $\mathcal{M}_T$ to be a binary classifier whose goal is to infer if $Y>c$, where $c$ is a task-specific threshold such as the average or the median value. Our attack can, however, be easily generalized to multi-class classification and regression tasks.

Our adversary aims to infer the Pearson correlation coefficient $\rho(X_1, X_2)$
between two input variables $X_1$ and $X_2$ given access to the machine learning model $\mathcal{M}_T$. 
Also called \textit{linear correlation}, the Pearson correlation coefficient measures the strength of the linear association between two variables and is widely used by researchers and data scientists across many domains \cite{waldmann2019use,zhou2016new}.
We call $\rho(X_1, X_2)$ the \textit{target correlation} and $X_1$ and $X_2$ the \textit{target variables}.

We make two assumptions on the adversary knowledge on the data distribution:

\begin{enumerate}
\item \textbf{Knowledge of one-way marginals.} First, we assume the adversary to have knowledge of the one-way marginals of all the variables $F_i(x), i=1, \ldots, n-1$ and $F_n(y)$. 
These are indeed typically released, e.g., in research papers or to comply with legal requirements such as the European Union AI Act~\cite{european2020artificial} and their knowledge is a common assumption in the literature~\cite{fredrikson2014privacy,fredrikson2015model,zhang2021leakage}. 

\item \textbf{Knowledge of some correlations.} Second, we assume the adversary to have access to the linear correlations between some of the variables. 
More specifically, we define the \textit{partial knowledge} of the adversary as a set $P$ of pairs such that for every $(i, j) \in P$ the adversary has the knowledge of the ground-truth values $\rho(X_i, X_j)$ (with $X_n=Y$). 
As $\rho(X_i, X_j)=\rho(X_j, X_i)$, by symmetry we assume that $P$ consists of ordered pairs, i.e., if $(i, j) \in P$, then necessarily $i<j$.

We consider three different attack scenarios: 
\begin{enumerate}
\itemsep0em
    \item[\textbf{(S1)}] The adversary knows the correlations between the target variables and the output variable, namely $\rho(X_1,Y)$ and $\rho(X_2,Y)$, i.e., $P= \{(1,n),(2,n)\}$, but not $\rho(X_i, X_j)$ for $1\leq i < j \leq n-1$ nor $\rho(X_i, Y)$ for $i \geq 3$.
    \item[\textbf{(S2)}] The adversary knows the correlations between all the input variables and the output variable: $(\rho(X_i, Y))_{i=1,\ldots,n-1}$, i.e., $P = \{(i,n ) \; : \; i = 1, \ldots, n-1 \}$, but the adversary does not know $\rho(X_i, X_j)$ for $1\leq i < j \leq n-1$.
    \item[\textbf{(S3)}] The adversary knows all the correlations between the variables, except for the target  correlation $\rho(X_1, X_2)$. More specifically, the adversary knows  $\rho(X_i, X_j), \forall 1 \leq i < j \leq n$ such that $(i, j) \neq (1, 2)$, i.e., $P = \{(i,j) \;:\; 1 \leq i < j \leq n\}\setminus \{(1,2)\}$.
\end{enumerate}
Note that the three scenarios are identical for $n=3$ variables.

We will henceforth call the known coefficients the \textit{correlation constraints}.

$\textbf{S2}$ is the default scenario considered in this paper. This choice is motivated by the fact that the relationship between the input and output variables -- encoded through $\rho(X_i, Y)$ -- is unlikely to be considered a secret, since the model's goal is to infer $Y$ given $X$. We furthermore show in Sec.~\ref{sec:discussion} how these correlation constraints can be extracted from the model.

$\textbf{S1}$ is a weak adversary, motivated by our goal to measure the \textit{leakage from the model}, rather than the leakage derived from constraints imposed on $X_1$ and $X_2$ via their association with the output variable $Y$. 
Indeed, our analysis shows that the correlation coefficients $\rho(X_1, Y)$ and $\rho(X_2, Y)$ restrict the range of values that the target coefficient $\rho(X_1, X_2)$ can take, without the adversary even accessing the model.

$\textbf{S3}$ is the strongest, ``worst-case'' scenario, corresponding to an adversary having complete knowledge of the correlations, except for the target. 
We mostly study this scenario in order to evaluate how much information a very strong adversary would be able to infer.

\end{enumerate}

We consider two different assumptions on the adversary access to the target machine learning model $\mathcal{M}_T$. 

\begin{enumerate}
\item \textbf{Model-less attack.} Under this attack scenario, the adversary does not have access to the target model $\mathcal{M}_T$. 
This scenario mostly serves as a stronger than random baseline allowing to measure the adversary's uncertainty on the target correlation under the prior knowledge. 
It is necessary to consider this scenario properly evaluate the leakage from the model.
A method for this scenario might also become an attack in situations where the model is not released, yet information about the marginals and the correlations between the input variables and the target variables is made available, e.g., as part of a research study. 

\item \textbf{Model-based attack.} Under this attack scenario, we assume the adversary 
(1) knows the model architecture and training details, e.g., number of training epochs for a neural network, allowing them to train from scratch a similar model and 
(2) has black-box query access to the target model $\mathcal{M}_T$, allowing to retrieve the output probabilities for each class $\mathcal{M}_T(x)$ on inputs $x$.
\end{enumerate}

\subsection{Attack methodology}
\label{subsec:results:attack-method}
\textbf{Overview.} We frame the \textit{correlation inference} task as a classification task by dividing the range of correlations $[-1, 1]$ into $N_B$ bins of equal length. The adversary's goal is to infer the correlation bin to which $\rho(X_1,X_2)$ belongs.  When $N_B=3$, the bins are $[-1, -1/3)$, $[-1/3, 1/3)$ and $[1/3, 1]$, which we refer to as \textit{negative}, \textit{low} and \textit{positive} correlations, respectively. 
We opt for a classification rather than a regression task  because it is easier to analyze the attack performance by comparing its accuracy with a random guess baseline (at $1/N_B$). Unless otherwise specified, we use $N_B=3$ classification bins in our empirical evaluation, as the correlation ranges can be easily interpreted. We also discuss in Sec.~\ref{subsec:results:increasing_n} and Sec.~\ref{subsec:results:real-world-datasets} results of our attack for $N_B=5$ bins.

At a high level, our methodology analyzes the behavior of the target model through the confidence scores returned on specific inputs, searching for patterns allowing to infer the correlations between input variables. 
More specifically, we train a meta-classifier to infer the correlation $\rho(X_1, X_2)$ between two variables of interest in a dataset $D$ based on features extracted from a model $\mathcal{M}$ trained on $D$.
Training the meta-classifier requires a large number of models $\mathcal{M}$ derived from datasets $D$ spanning a wide variety of values for the unknown correlations $\rho(X_i, X_j)$, where $(i, j) \notin P$, while matching the correlation constraints $\rho(X_i, X_j)$ for every $(i, j) \in P$. 
Generating datasets satisfying this requirement on their correlations is a key technical challenge our work addresses.

\textbf{Correlation matrices.} To formalize this requirement on the linear relationships between the $n$ variables in a dataset, we use the concept of \textit{correlation matrix}.
This is the $n \times n$ real-valued matrix $C$ of Pearson correlation coefficients between all the pairs of variables: $c_{i,j} = \rho(X_i, X_j), i,j=1,\ldots,n$. 
Note that we use lowercase notation to denote matrix coefficients. 
Let $\mathcal{C}$ be the set of \textit{valid correlation matrices}.
$\mathcal{C}$ consists of all $n\times n$ real-valued matrices satisfying the following properties:

\begin{enumerate}
    \itemsep0em 
    \item[\textbf{(P1)}] All elements are valid correlations, i.e., real values between $-1$ and $1$: $-1 \leq c_{i,j} \leq 1, \forall 1 \leq i, j \leq n$.
    \item[\textbf{(P2)}] Diagonal entries are equal to $1$, i.e., there is perfect correlation between a variable and itself: $c_{i,i} = 1, \forall 1 \leq i \leq n$.
    \item[\textbf{(P3)}] The matrix is symmetric, i.e., the correlation between $X_i$ and $X_j$ is the same as the correlation between $X_j$ and $X_i$: $c_{i,j}=c_{j,i}$. 
    \item[\textbf{(P4)}] The matrix is positive semi-definite: $x^T C x \geq 0, \forall x \in \mathbb{R}^n$. 
\end{enumerate}

The \textbf{spherical parametrization} of correlation matrices~\cite{pinheiro1996unconstrained} provides a principled and effective approach for sampling valid correlation matrices $C$. 
This approach crucially relies on the Cholesky decomposition of $C=BB^T$ which, as we describe next, can be used to generate valid correlation matrices by sampling its correlation coefficients one-by-one, conditionally on the correlation constraints.
More specifically, since $C$ is positive semi-definite, it has a Cholesky decomposition $C=B B^T$, where $B$ is lower triangular with non-negative diagonal entries: $b_{i, j}=0, 1 \leq i < j \leq n$ and $b_{i, i} \geq 0, i=1,\ldots, n$.
Furthermore, the coefficients of matrix $B$ can be expressed using spherical coordinates, as follows: 
{
\begin{equation}
   \begin{aligned} 
        B = \begin{pmatrix}
        1  & 0 & 0 & \cdots &  0\\
        \cos\theta_{21} & \sin\theta_{21} &  \ddots & \ddots & \vdots \\
        \cos\theta_{31}  & \cos\theta_{32}\sin\theta_{31} & \sin\theta_{32}\sin\theta_{31} & \ddots & \vdots\\
        \vdots & \vdots & \vdots & \ddots  & \vdots \\
        \cos\theta_{n1}& \cos\theta_{n2}\sin\theta_{n1} & \cos\theta_{n3}\sin\theta_{n2}\cdot \sin\theta_{n1} & \cdots &   \prod\limits_{j-1}^{n-1}\sin\theta_{nj}.
        \end{pmatrix}
    \end{aligned}
\end{equation}
}
This is because the diagonal of $C$ consists of ones, i.e., $1= c_{i,i} = (B B^T)_{i, i} $ $=\sum_{j=1}^n b_{i,j}^2=\sum_{j=1}^i b_{i,j}^2$, which means that each vector $(b_{i,1}, \ldots, b_{i, i}) \in \mathbb{R}^i$ has an $\mathbb{L}_2$ norm of 1 and can therefore be expressed using spherical coordinates, i.e., $\exists \theta_{i,j}\in [0, \pi), 1 \leq i < j < n$.
Note that as expected for the symmetric matrix $C$, only $\frac{n \times (n-1)}{2}$ parameters, called angles, are sufficient to characterize the set of valid correlation matrices. 

A key property exploited by both our model-less and model-based attacks is that the first column of $B$ is identical to the first column of $C$. By reordering--without loss of generality--the variables as $Y, X_1, \ldots, X_{n-1}$, all the correlation constraints appear on the first column as free parameters under scenarios \textbf{S1} and \textbf{S2}.
We will use this property to sample the other correlation coefficients conditionally on the correlation constraints (Alg. S2 and ~\ref{alg:sample_correlation_matrix_s2}).
We will further show how sampling can be done even in scenario \textbf{S3} (Alg.~\ref{alg:sample_correlation_matrix_s3}), where the correlation constraints cannot be expressed using only the free parameters of the first column.
We start by describing the model-less attack, to build intuition into how this property is useful for sampling correlation matrices under constraints.

\textbf{Model-less attack.} Our \textit{model-less attack} leverages the intuition that the adversary's knowledge of some correlations allows to derive theoretical limits on the range spanned by the other values.
This allows the adversary to make a better than random guess about the target correlation $\rho(X_1, X_2)$ without access to the model.
The model-less attack acts as a strong baseline for any attack using access to the model, allowing to correctly quantify the leakage caused by the model.

Given the correlation constraints $(c_{i, j})_{(i, j) \in P}$,
we call \textit{matching set} $C_P$ the set of valid correlation matrices satisfying them: 
$
C_P = \{ C' \in \mathcal{C} \; : \; \forall (k, l) \in P \;, \; c'_{k, l} = c_{k, l} \}
$.
Due to the correlation constraints, the range of values $R$ that can be taken by $c_{1,2}$ is a subinterval of $[-1, 1]$:  $R= [\inf_{C' \in C_P} c'_{1,2},\sup_{C' \in C_P} c'_{1,2}] := [m_1, m_2]$.
Our model-less attack outputs the majority bin over this interval. 
To avoid introducing biases in the evaluation, we assume the uniform prior over this interval and return the classification bin which is covered in the largest proportion by $[m_1, m_2]$ (see Supplementary Materials ~\ref{appendix:details_model_less_attack} for details).

As an example, consider the case $n=3$ and an adversary having knowledge of $\rho(X_1, Y)$ and $\rho(X_2, Y)$ (corresponding to $P = \{(1,2), (1,3)\}$).
We assume, for simplicity, that the correlation matrix is computed over $Y, X_1, X_2$ in this order.
Developing the computation $C=B B^T$ yields the following correlation coefficients: 
\begin{align}
    c_{2,1} &= \cos~\theta_{2,1} = \rho(X_1, Y)\notag \\
    c_{3,1} &= \cos~\theta_{3,1} = \rho(X_2, Y)\label{eq:coef32} \\
    c_{3,2} &= \cos~\theta_{2,1} \cos~\theta_{3,1} + \sin~\theta_{2,1}\sin~\theta_{3,1}\cos~\theta_{3,2} = \rho(X_1, X_2) \notag
\end{align} 
Since $\theta_{2,1}$ and $\theta_{3,1}$ are fixed, the only degree of freedom left on the unknown $\rho(X_1, X_2)$ is $\theta_{3,2}$. 
Any choice of $\theta_{3,2} \in [0, \pi)$ yields a valid correlation matrix.
The unknown $\rho(X_1, X_2)$ can thus take any value between $\cos~\theta_{2,1} \cos~\theta_{3,1} - \sin~\theta_{2,1}\sin~\theta_{3,1}=\cos(\theta_{2,1}+\theta_{3,1})$ and  $\cos~\theta_{2,1} \cos~\theta_{3,1} + \sin~\theta_{2,1}\sin~\theta_{3,1} = \cos(\theta_{2,1}-\theta_{3,1})$, but cannot fall outside the interval determined by these endpoints.
Depending on the values of the constraints, this interval might be much smaller than $[-1, 1]$, thereby reducing the uncertainty of the adversary. 

\begin{figure*}[!ht]
\centerline{\includegraphics[width=0.65\linewidth]{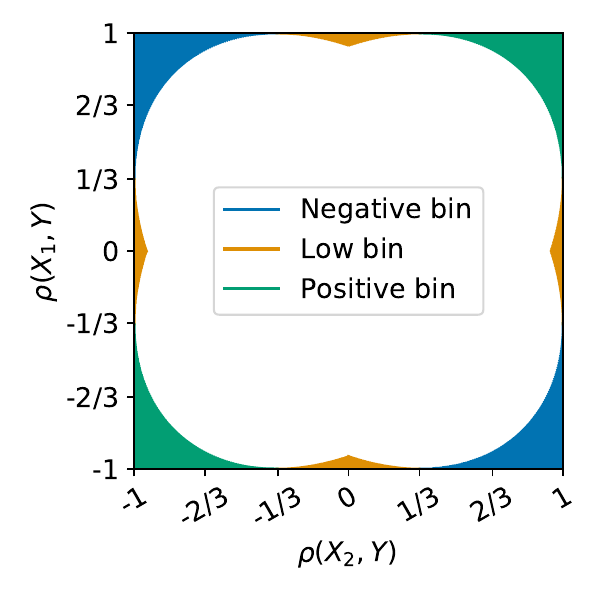}}
\caption{\textbf{Illustration of correlation constraints $(\rho(X_1, Y), \rho(X_2, Y))$ that lead to only one possible bin for the unknown correlation $\rho(X_1, X_2)$.} Here, we show results for $N_B=3$ possible bins.}
\label{fig:ear-figure}
\end{figure*}

Fig.~\ref{fig:ear-figure} shows that some constraints yield a theoretical interval $R$ which is strictly included in one of the $N_B=3$ classification bins. 
For such constraints, the adversary can infer the target correlation bin with certainty.
These constraints are of the two types. 
First, constraints that are close to 1 in absolute value result in high $\rho(X_1, X_2)$ belonging to the positive, green bin, when they have the same sign and to the negative, blue bin, when they have opposite signs. 
Second, when one of the constraints is close to 0 while the other is close to 1 in absolute value, $\rho(X_1, X_2)$ is necessarily in the low, orange bin. 
We refer the reader to Supplementary Materials~\ref{appendix:shape_analysis} for a precise, mathematical description of these regions.

This interval can be computed analytically or empirically, by sampling a sufficient number of correlation matrices satisfying the correlation constraints. 
We here use the latter approach, as our model-based attack also requires generating valid correlation matrices satisfying the constraints given by the adversary knowledge. 
We refer the reader to Appendix~\ref{appendix:model-less-number-of-samples} for an analysis of the impact of the number of samples on the model-less attack performance.

\textbf{Generating correlation matrices under constraints.} To generate valid correlation matrices satisfying the correlation constraints, we adapt an algorithm developed by Numpacharoen and Atsawarungruangkit~\cite{numpacharoen2012generating}. 
The goal of the original algorithm, which we include in Supplementary Materials~\ref{appendix:sample_correlation_matrix} (see Alg.~\ref{alg:sample_correlation_matrix}),  is to sample valid correlation matrices (under no constraints).
The insight of the authors is that each correlation coefficients $c_{i,j}$ can be expressed as $c_{i, j} = m_{i,j} + \cos\theta_{i, j} l_{i, j}$, where $m_{i, j}$ and $l_{i,j}$ only depend on $\theta_{p,q}$ for $1 \leq p \leq i$ and $1 \leq q < j$. 
Alg.~\ref{alg:sample_correlation_matrix} uses this insight to generate valid correlation matrices by sampling the correlations in order, from top to bottom ($i=1, \ldots, n$) and from left to right ($j=1, \ldots, n$), uniformly within the bounds $m_{i,j } \pm l_{i, j}$ derived from the values previously sampled. 
The formulas on $m_{i,j}$ and $l_{i, j}$ can be derived (supposing without loss of generality that $i \geq j$) by developing the computation of $c_{i,j}=(B B^T)_{i,j}=B_{i,1:j-1}(B_{j,1:j-1})^T + b_{i,j}b_{j,j}$ and setting $m_{i,j}=\sum_{p=1}^{j-1} (\cos~\theta_{i,p}\cos~\theta_{j,p}\prod_{q=1}^{p-1}\sin~\theta_{i,q}\sin~\theta_{j,q})$ and $l_{i,j}= \prod_{q=1}^{j-1}\sin~\theta_{i,q}\sin~\theta_{j,q}$. 

We adapt this algorithm to our problem by noticing that the coefficients of the first column of $B$ ($\cos~\theta_{i, 1}$ for  $i=2, \ldots, n-1$) are identical to those of $C$ and are free parameters. 
We set these coefficients equal to the correlation constraints and sample the others uniformly within their bounds.
We propose two adaptations of the original algorithm for attack scenarios \textbf{S1} and \textbf{S2}, Alg.S2 and Alg.~\ref{alg:sample_correlation_matrix_s2}, respectively.
Note that correctly sampling the correlation matrices without biasing the values requires carefully re-ordering the variables before and after the sampling, an intricate operation that we detail in Sec.~\ref{subsec:materials-methods:generate-correlation-matrices}.
We further propose an algorithm to sample the unknown value $\rho(X_1, X_2)$ for attack scenario \textbf{S3}, when all the other correlations are known (Alg.~\ref{alg:sample_correlation_matrix_s3}).

\textbf{Model-based attack.}
We finally develop a \textit{model-based attack}, which trains a meta-classifier to infer the bin to which $\rho(X_1,X_2)$ belongs on a labelled dataset of model and correlation pairs ($M(D), \rho(X_1(D), X_2(D))$).
We create this labelled dataset by training models on synthetic datasets generated using the Gaussian copula generative model~\cite{sklar1959fonctions}.
We select the Gaussian copula model because it is parameterized by one-way marginals and a covariance matrix, allowing us to generate synthetic datasets that match the adversary's knowledge about the target dataset $D_T$, while spanning the
entire range of values attainable by the unknown correlations.
Using our notation, the Gaussian copula model is parameterized by (1) the marginals of the variables $(F_i)_{1\leq i \leq n}$
$~$and (2) a positive semi-definite covariance matrix $\Sigma$ yielding the following joint distribution: 
\begin{equation}\text{Pr}(X_1 \leq x_1, \ldots, X_n \leq x_n) = \Phi_\Sigma \big( \Phi^{-1}(F_1(x_1)), \ldots, \Phi^{-1}(F_n(x_n)) \big),
\end{equation}
where $\Phi_\Sigma$ is the cumulative distribution function (CDF) of the Gaussian multivariate distribution $\mathcal{N}(0, \Sigma)$: $
\Phi_{\Sigma}(x_1, \ldots, x_n) = \bigints_{-\infty}^{x_1} \ldots \bigints_{-\infty}^{x_n} \frac{e^{-\frac{1}{2}x^T \Sigma^{-1} x}}{\sqrt{(2\pi)^n \det(\Sigma)}}$
and $\Phi$ is the CDF of a one-dimensional standard normal $\mathcal{N}(0, 1)$. Alg.~\ref{alg:sample_from_gaussian_copulas} in  Supplementary Materials~\ref{appendix:sample_from_gaussian_copulas} describes the procedure to sample from the Gaussian copulas.

Consistent with the rest of our methodology, we use correlation, rather than covariance matrices to parametrize the Gaussian copulas. Correlation matrices are a subset of covariance matrices that have only values of 1 on the diagonal.

To generate synthetic datasets $D$ matching the correlation constraints and the one-way marginals of the private dataset $D_T$, we distinguish between two cases:

\begin{enumerate}
\item When the variables are all standard normals, i.e., $F_i = \Phi, i=1, \ldots, n$, the correlations of the distribution output by the Gaussian copula are identical to the parameter $\Sigma$. 
We first generate a correlation matrix $C$ using the algorithm corresponding to the attack scenario (e.g., Alg.~\ref{alg:sample_correlation_matrix_s2} for \textbf{S2}), then set $\Sigma=C$ to sample a dataset whose empirical correlation approximately match $C$.

\item When the variables have arbitrary one-way marginals, the correlations of the distribution output by the Gaussian copula parameterized by these marginals are not necessarily equal to the parameter $\Sigma$~\cite{xiao2016calculating}.
This means that the adversary must not directly parameterize the Gaussian copula by setting $\Sigma=C$, but instead \textit{must modify the correlation constraints} $V = \{ (c_T)_{i,j}, (i,j) \in P \}$ (with $C_T$ denoting the correlation matrix of private dataset $D_T$) to a new set of constraints $V'$, generate a correlation matrix $C'$ under the constraints given by $V'$ (instead of $V$), and only then set $\Sigma=C'$.
$V'$ is chosen such that the empirical correlations computed on the synthetic datasets are approximately equal to $V$ and thus match the correlation constraints.
Alg.~\ref{alg:shift_correlation_constraints} details a simple, yet very effective heuristic to modify the correlation constraints.

\end{enumerate}

To train our meta-classifier, we first generate $K$ synthetic datasets $D^k, k=1,\ldots, K$.
Second, we train a model $\mathcal{M}^k$ on each dataset using the same architecture and hyperparameters as the target model, but a different seed (we refer the reader to Sec.~\ref{sec:discussion} for results of our attack when the adversary knows the seed).
Finally, we extract a set of features from each model in the form of confidence scores computed on a synthetic query dataset $D_{\text{query}}$ generated in the same way as the other datasets, i.e., to match the adversary's knowledge.
We train the meta-classifier to infer the correlation $\rho(X_1(D^k), X_2(D^k))$, discretized over $B$ bins, given as input the features extracted from the model $\mathcal{M}^k$.
The complete procedure is described in Sec.~\ref{subsec:materials-methods:model-based-attack}.

\subsection{Empirical evaluation}
\label{subsec:results}
We apply our attack to Logistic Regression (LR) and Multilayer Perceptron (MLP) models, which are standard choices for learning tasks on tabular training datasets.
We train the models on synthetic and real-world datasets, as described in the following sections.

Unless otherwise specified, we assume the adversary to have black-box access to a target model $\mathcal{M}_T$ trained on a target dataset $D_T$, with the goal of inferring the correlation $\rho(X_1,X_2)$.
Unless otherwise specified, we consider the default attacker \textbf{(S2)}, i.e., the attacker knows $\rho(X_i, Y)_{i=1,\ldots,n-1}$.

Across experiments, we run our correlation inference attacks against the target model by, first, training a meta-classifier on features extracted from models trained on synthetic datasets satisfying the adversary's knowledge about the target dataset $D_T$.
Second, we apply the meta-classifier to features extracted from the target model $\mathcal{M}_T$ to infer the bin to which the target correlation $\rho(X_1, X_2)$ belongs.
We repeat the experiment multiple times and compute the attack accuracy as the fraction of correct guesses.
Complete details of the experiments are provided in Supplementary Materials~\ref{appendix:experimental-setup}.

To quantify how much information is leaked by different types of models, we first compare the performances of our model-based and model-less attacks in a controlled setting.
More specifically, we train the target models $\mathcal{M}_T$ on  synthetic data generated using the Gaussian copula generative model.
Recall that this model is parameterized by a correlation matrix, which allows us to explore the entire range that can be taken by the correlation constraints $\rho(X_i, Y)_{i=1, \ldots,n-1}$ and characterize their impact on the performance of the attack. 

\subsubsection{Impact of constraints for $n=3$ variables}
\label{subsec:results:grid-evaluation}
To characterize the impact of the correlation constraints on the attack accuracy, we study a simple setting in which there are only two input variables and one output variable ($n=3$).
We consider all possible ranges for the correlation constraints $\rho(X_1, Y)$ and $\rho(X_2, Y)$ by dividing the grid $[-1,1]\times[-1,1]$ into $200\times200$ equal cells. 
We report one accuracy for every cell $[a_1,a_2]\times[b_1,b_2]$,  evaluating the performance of an attacker (model-less or model-based) at inferring the target correlation $\rho(X_1, X_2)$ for datasets subject to correlation constraints belonging to that cell, i.e., such that $a_1\leq\rho(X_1, Y)<a_2$ and $b_1\leq\rho(X_2, Y)<b_2$.

In every cell, we draw $T'$ pairs of constraints $\rho(X_1, Y)$ and $\rho(X_2, Y)$ uniformly in $[a_1,a_2]$ and $[b_1,b_2]$, respectively.
Each pair is used to generate, in order: (1) a target correlation matrix $C_T$ using Alg.~\ref{alg:sample_correlation_matrix_s2}, (2) a target dataset $D_T$ of $1000$ samples using Gaussian copulas parametrized by $C_T$ and standard normal one-way marginals, and (3) a target model $\mathcal{M}_T$ trained on $D_T$ to perform the binary classification task $Y>0$. 
We make three observations regarding (1).
First, since there are only $n=3$ variables, Alg.~\ref{alg:sample_correlation_matrix_s2} and Alg. S2 are equivalent. Second, $\rho(X_1, X_2)$ is the only correlation coefficient being generated. Third, this value is sampled uniformly over the range defined by Eq.~\ref{eq:coef32} and correlation constraints $\rho(X_1, Y)$ and $\rho(X_2, Y)$.
For every pair of constraints, given access to it and to $\mathcal{M}_T$, the adversary's goal is to infer $(c_T)_{1,2}=\rho(X_1, X_2)$.
Executing our attack independently on all the $T'$ targets in every cell would require us to generate $K\times T'\times200\times200$ models in order to train meta-classifiers, which is computationally infeasible.
Instead, in every cell, we (1) run the model-based attack by training one meta-classifier and (2) run the model-less attack by empirically estimating, from the samples, the range of values that $\rho(X_1,X_2)$ can take when the correlation constraints belong to that cell.
We evaluate the attacks using 5-fold cross-validation and setting $T'=1500$.

\begin{figure*}[!htbp]
\centerline{\includegraphics[width=\linewidth]{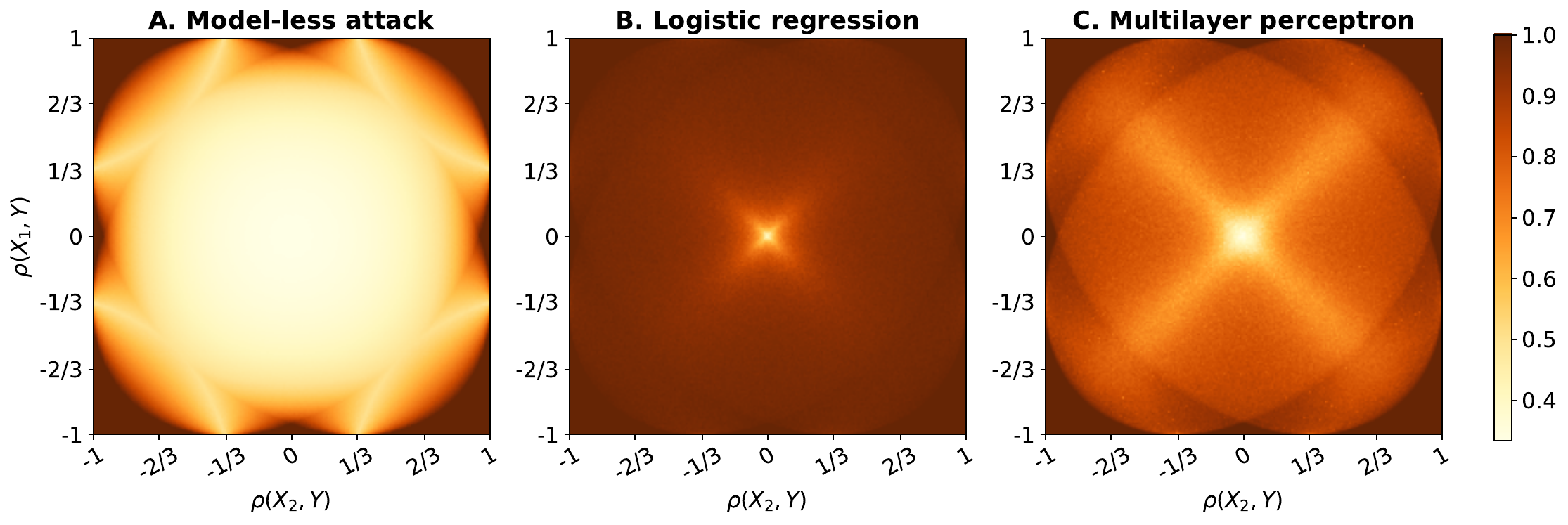}}
\caption{\textbf{Accuracy of the correlation inference attack on $n=3$ variables as a function of the correlation constraints $\rho(X_1, Y)$ and $\rho(X_2, Y)$.}  
We show results of (A) the model-less attack and of the model-based against two different target models: (B) logistic regression and (C) multilayer perceptron.
The color of each cell indicates the accuracy of our correlation inference attack applied to models trained on synthetic datasets whose correlations $\rho(X_1, Y)$ and $\rho(X_2, Y)$ belong to the region defined by the cell. There are $200\times200$ cells in total.
}
\label{fig:grid-attack}
\end{figure*}

Fig.~\ref{fig:grid-attack}A shows an adversary with no access to the model can infer the target correlation $\rho(X_1, X_2)$ 56.0\% of the time on average over all the cells, much better than random (33.3\%).
These results confirm the theoretical analysis presented in Sec.~\ref{subsec:results:attack-method} with our model-less attack achieving perfect accuracy in the corners of the grid and the top center region (with its symmetric counterparts).
They also highlight why it is essential to have a strong baseline different from the random guess baseline in order to quantify the leakage from the model.
The accuracy of the model-less attack then slowly decreases as the constraints move to regions of higher uncertainty, in which more than one bin is possible.

Fig.~\ref{fig:grid-attack}B shows that black-box access to a Logistic Regression (LR) allows an adversary to greatly improve the accuracy of the attack, reaching 95.6\% on average over all the cells. 
As expected, the model-based attack achieves perfect accuracy in the regions where the model-less attack also does.
Outside these regions, the model-based attack vastly outperforms the model-less attack.
Its accuracy exceeds 90\% everywhere, with the exception of the cross-like region centered in the origin where the models leak less information than elsewhere. 
At the center of the cross ($|\rho(X_i, Y)| \leq 0.02, i=1,2$), where the input variables $X_1$ and $X_2$ are uncorrelated with the output variable $Y$, the accuracy of our attack is 47.2\%.  
When there is little to no correlation between the inputs and the output variable $Y$ ($\rho(X_i, Y)\simeq0$), the target models do not learn much about the dataset. 
As for the arms of the cross, where $\rho(X_1,Y)\simeq\rho(X_2,Y) \ll 1$, we believe that the lower attack performance is due to the theoretical interval attainable by the target correlation $\rho(X_1, X_2)$ being close to $[-1,1]$ and hence the adversary having maximum uncertainty.
Even in this scenario, the leakage from the model is much higher than the baseline.

Fig.~\ref{fig:grid-attack}C shows that multilayer perceptron (MLP) models are also vulnerable to our attack, achieving an accuracy of 82.3\% on average over the cells. 
This shows that access to an MLP model greatly improves over the model-less attack, but slightly less than the LR.
This may seem surprising at first, since our MLP, with its 292 learnable weights, has a much higher capacity than the LR with its 3 weights and thus a higher ability to retain information about the dataset. 
We believe that this gap is due to the fact that the non-convex loss used to train MLP models has a (potentially large) number of local minima~\cite{choromanska2015loss}.
This means that the models trained by an adversary's in order to generate the meta-classifier's training dataset are unlikely to reach the same local minima as the target model, increasing the adversary's uncertainty. 
We refer the reader to Sec.~\ref{sec:discussion} for results of our attack when we remove the uncertainty caused by the difference in randomness. 
We also find that the cross-like behavior is exacerbated on the MLP: the center of the cross is larger, reaching an average accuracy of 48.1\% for $|\rho(X_1, Y)| \leq 0.1$.

\textbf{White-box results.} Fig. \ref{fig:grid-attack-logreg} and ~\ref{fig:grid-attack-mlp} in the Supplementary Materials show that white-box access to the model, where the features extracted from the model include its weights, does not improve the performance of our attack. 
We refer the reader to the Supplementary Materials~\ref{appendix:white_box_results} in for details of the experiment and an analysis of the results.

\subsubsection{Increasing the number of variables $n$}
\label{subsec:results:increasing_n}

Next, we compare the performances of our attack against Logistic Regression and MLP models when increasing the number of variables $n$, under three different attack scenarios.

For each $n=3, \ldots, 10$, we sample 1000 target correlation matrices $C_T\in\mathcal{C}$ using Alg.~\ref{alg:sample_correlation_matrix}.
To obtain a balanced distribution over  the target correlations, we generate the value $\rho(X_1, X_2)$ first, as explained in Sec.~\ref{subsec:materials-methods:generate-correlation-matrices}.
Then for each $C_T$, we sample a synthetic target dataset $D_T$ (parametrized by $C_T$ and standard normal marginals) and train a target model $\mathcal{M}_T$ on it.

\begin{figure}[h]
\centerline{\includegraphics[width=\linewidth]{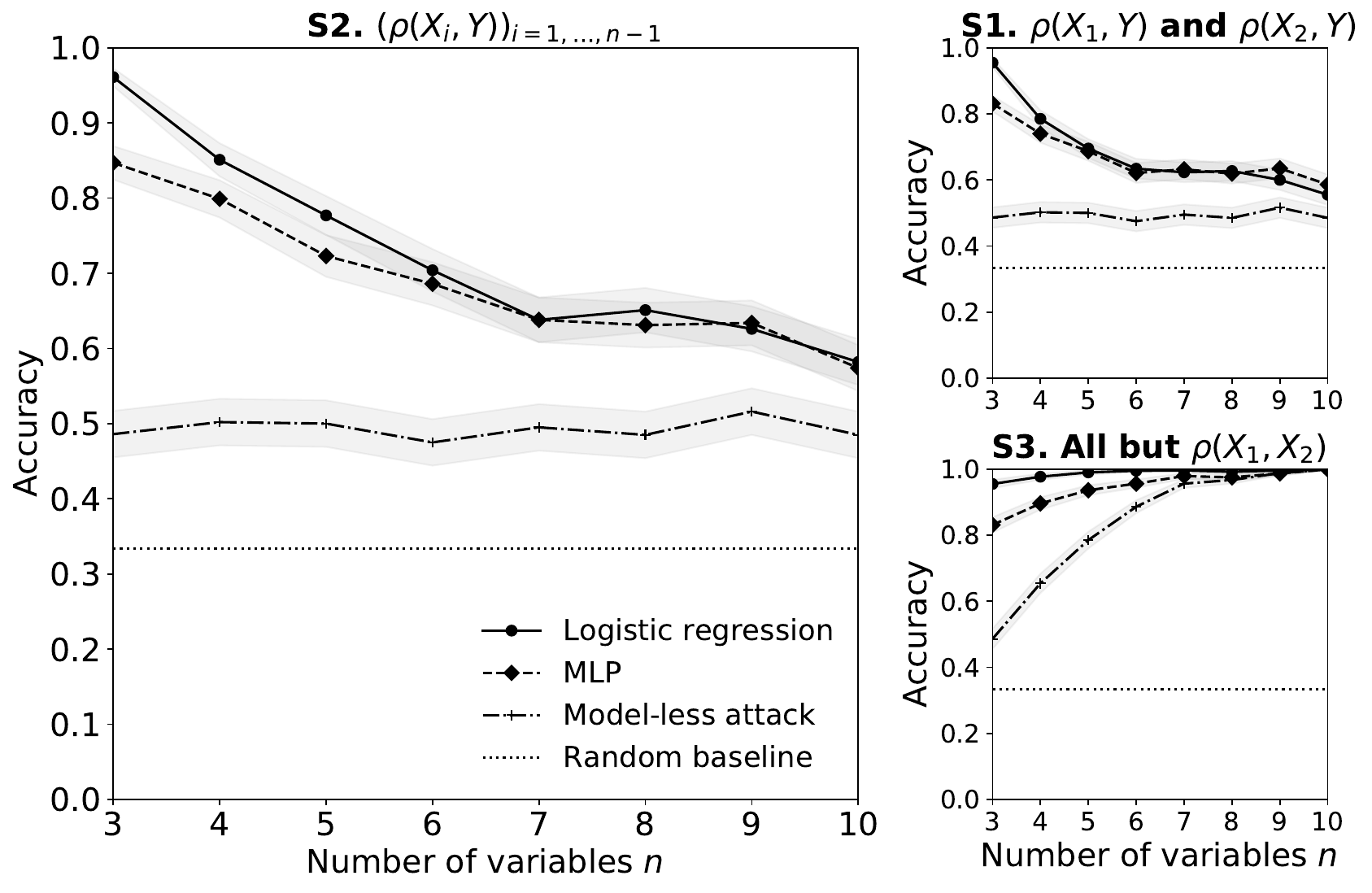}}
\caption{\textbf{Attack accuracy for different scenarios and number of variables in the dataset $n$}. We consider three different attack scenarios: \textbf{(S1)} the adversary knows the correlations between the target variables and the output variable $\rho(X_1,Y)$ and $\rho(X_2,Y)$, \textbf{(S2)} the adversary knows the correlations between all the input variables and the output variable $\rho(X_i,Y)_{i=1,\ldots,n-1}$, and \textbf{(S3)} the adversary knows all the correlations between the variables except for the target correlation $\rho(X_1,X_2)$. We compute the accuracy (with 95\% confidence interval) over 1000 target models.}
\label{fig:rand-tar-attack-comparison-constraints}
\end{figure} 

Fig.~\ref{fig:rand-tar-attack-comparison-constraints} \textbf{S2} shows that the adversary is consistently and significantly better than the model-less attack.
The performance of our attack decreases slowly with the number of variables in the dataset, from 96.1\% ($n=3$) to 70.4\% ($n=6$) and 58.2\% ($n=10$) when applied to LR models, while the model-less attack has a roughly constant accuracy of 49.3\% for all values of $n$.
This decrease when $n$ increases is expected. 
Indeed, first, the number of unknown correlations increases quadratically with $n$, which increases the uncertainty of the adversary. 
Second, the more variables a model is trained on, the less likely our target variables $X_1$ and $X_2$ are to be useful for the prediction task at hand. 
This in turn means the model might learn less information about $\rho(X_1, X_2)$.

Fig.~\ref{fig:impact-of-largest-constraint} in the Supplementary Materials shows that when at least one of the constraints $\rho(X_1, Y)$ or $\rho(X_2, Y)$ is large in absolute value (meaning that $X_1$, $X_2$, or both are good predictors of the output variable), our attack performs better.
For instance, on $n=6$ variables our attack reaches 90.6\% when $\max(|\rho(X_1, Y)|, |\rho(X_2, Y)|)\geq0.8$, much better than 62.7\% when $\max(|\rho(X_1, Y)|, |\rho(X_2, Y)|)\leq0.2$.
Interestingly, even when both constraints are small, e.g.,  when $\max(|\rho(X_1, Y)|, |\rho(X_2, Y)|)\leq0.2$, the model-based attack can still infer $\rho(X_1, X_2)$ much better than random, especially for a smaller number of variables.
We discuss potential reasons for this behavior in Sec.~\ref{sec:discussion}.

Interestingly, the difference in vulnerability between Logistic Regression and MLP models reduces as the number of variables in the dataset $n$ increases, with our attack achieving similar accuracy against the two for $n\geq6$.
We rule out the possibility of models learning the same decision boundary, as the MLP models are, on average, 1 to 4\% less accurate, with the gap in accuracy increasing with $n$.
We believe this could instead be due to MLP models' training algorithm becoming more stable as $n$ increases, i.e., the loss having fewer local minima, as more input variables are likely to be predictive of the output $Y$.

We obtain similar trends on $N_B=5$ classification bins, and refer the reader to Fig.~\ref{fig:rand-tar-attack-5-bins} in the Supplementary Materials for detailed results.

Fig.~\ref{fig:rand-tar-attack-comparison-constraints} \textbf{S1} shows that reducing the knowledge of correlations to only $\rho(X_1, Y)$ and $\rho(X_2, Y)$, only slightly decreases the attack accuracy.
More specifically, the accuracy is 78.5\% when $n=3$ vs 85.1\% for \textbf{S2}, 63.4\% when $n=6$ vs. 70.4\% for \textbf{S2}, and 55.5\% when $n=10$ vs. 58.2\% for \textbf{S2}.
This shows that correlation inference attacks are a concern even when facing a weak adversary.

We complete our understanding of the impact of uncertainty on the attack by additionally considering a very strong adversary, who has knowledge of all the correlations but the target $\rho(X_1, X_2)$.
Fig.~\ref{fig:rand-tar-attack-comparison-constraints} \textbf{S3} shows that the accuracy of our attack now increases as $n$ increases.
We believe this is because the fraction of unknown correlations now reduces quadratically with $n$.
More specifically, our attack improves as $n$ increases, from 95.5\% ($n=3$) to 99.5\% ($n=6$) and 99.9\% ($n=10$) on Logistic Regression models.
We obtain similar results on MLP models, as our attack improves from 83.1\% ($n=3$) to 95.6\% ($n=6$) and 99.9\% ($n=10$).
Interestingly, the strongest adversary can again exploit a higher vulnerability in LR models compared to MLP models.
The model-less attack also increases with $n$, but achieves a lower accuracy than our model-based attack for $n\leq8$.

\subsubsection{Impact of mitigations}
\label{subsec:results:mitigations}

We analyze the effectiveness of three possible mitigations in preventing our correlation inference attack.

\begin{figure}[!htbp]
    \centering
    \includegraphics[width=\linewidth]{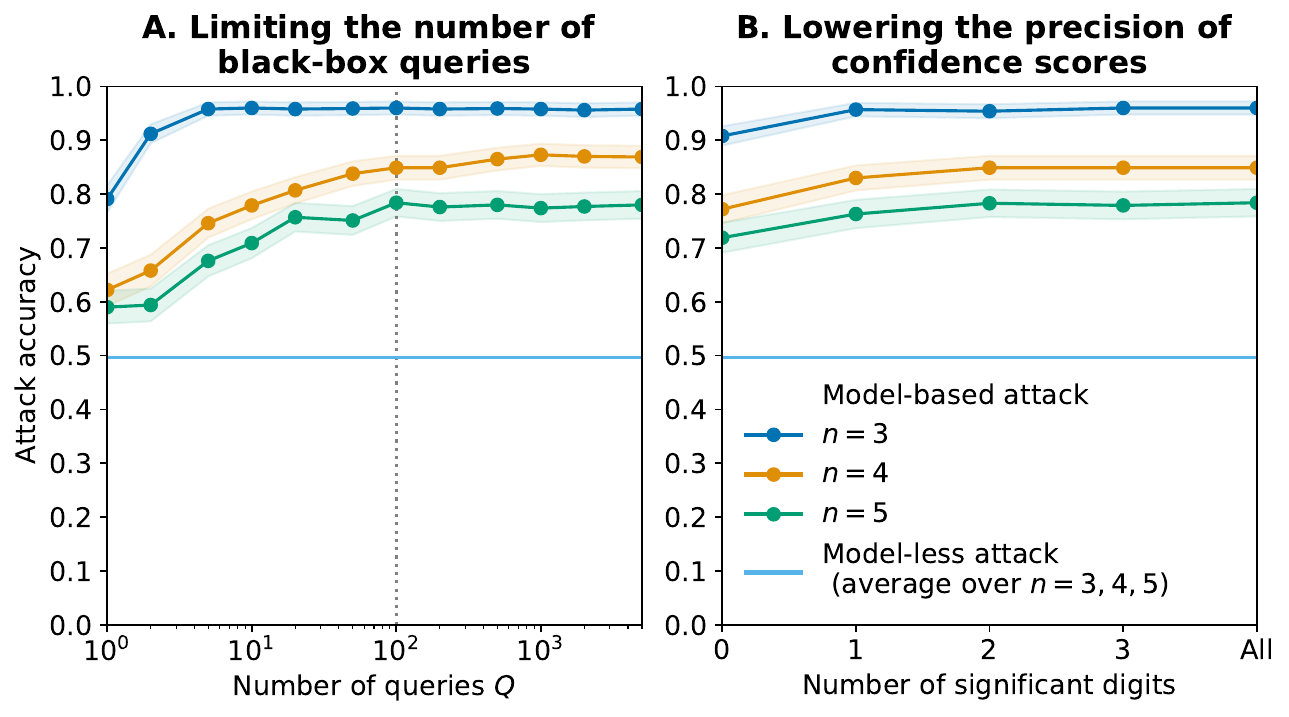}
    \caption{\textbf{Impact of mitigations on the accuracy of our attack against Logistic regression models.} We report results for two mitigations: limiting the number of black-box queries (A) and lowering the precision of confidence scores (B). We report the accuracy  (with 95\% confidence interval) of our model-less and model-based attacks over 1000 targets models for $n \in \{3,4,5\}$ variables.}
    \label{fig:mitigations-logreg}
\end{figure}

\textbf{Limiting the number of queries} could be put in place, e.g. in the machine learning as a service case, by the model developer in order to prevent the user from extracting too much information.
We vary the number of queries $Q$ made to an LR from 1 to 5000 and report the attack accuracy against the same targets as before.
Fig.~\ref{fig:mitigations-logreg}A shows that this measure does not prevent our attack.
Indeed only 5 queries are necessary to reach close to optimal accuracy for $n=3$, 50 queries for $n=4$ and 100 for $n=5$. Even a single query is sufficient for the attack to reach 79.1\% on $n=3$ variables, 62.2\% on $n=4$, and 59.0\% on $n=5$, significantly higher than the model-less attack which only reaches 49.6\%. 

\textbf{Lowering the precision of confidence scores} is another popular mitigation deployed when releasing models, mainly by allowing the model to only output the class label~\cite{choquette2021label}.
Fig.~\ref{fig:mitigations-logreg}B shows that this measure has a very small impact on our attack, only decreasing its accuracy from 96.0\% when all digits are made available to 90.8\% when only the label is, on $n=3$ variables (similarly, from 84.9\% to 77.2\% when $n=4$ and from 78.4\% to 71.9\% when $n=5$). We obtain similar results on MLP models, as shown on Fig.~\ref{fig:mitigations-mlp} in the Supplementary Materials.

\begin{figure}[!htbp]
    \centering
    \includegraphics[width=\linewidth]{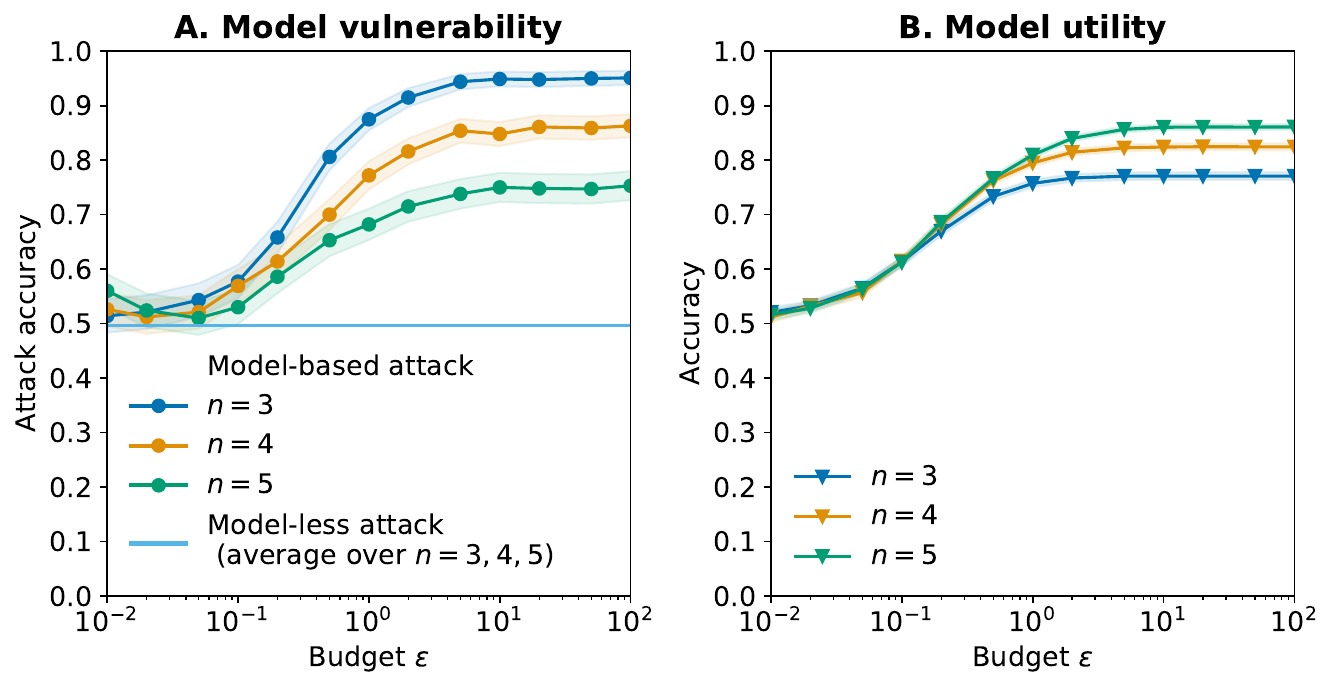}
    \caption{\textbf{Trade-off between the vulnerability of DP Logistic regression models to our attack (A) and their utility (B).} We compute the metrics (with 95\% confidence interval) over 1000 target models for $n \in \{3,4,5\}$ variables.
    }
    \label{fig:mitigations-dp}
\end{figure}

\textbf{Differential privacy (DP)} is a popular privacy definition requiring that the model should not depend too strongly on any one record, as controlled by $\epsilon$~\cite{dwork2006calibrating}. DP protects individual privacy and ML models trained with DP guarantees have been shown to prevent individual-level attacks, including membership inference attacks.
Although DP is not meant to protect population-level information such as correlations, the noise added might perturb the model enough to mitigate our attack.
We train 1000 DP Logistic Regression models with objective perturbation~\cite{chaudhuri2011differentially} using the Diffprivlib library~\cite{diffprivlib} and different values of $\epsilon$. 

Fig.~\ref{fig:mitigations-dp} shows that the vulnerability of a model to our attack is strongly linked to its utility, i.e., accuracy. 
Indeed, for large value of $\epsilon$, models are both useful and vulnerable. 
As we decrease $\epsilon$, more noise is added which reduces both the vulnerability of models to our attack but also their utility. 
For small value of $\epsilon$, our attack does not perform better than the model-less one, but at the cost of models losing any utility.

Taken together, our results suggests these mitigations to be ineffective in preventing our attack,

\subsubsection{Real-world dataset evaluation}
\label{subsec:results:real-world-datasets}
We evaluate the performance of our attack on three real-world tabular datasets: Fifa19~\cite{fifa192019}, Communities and Crime~\cite{communitiesandcrime2011}, and Musk~\cite{muskv21998} (see Supplementary Materials~\ref{appendix:experimental-setup:datasets} for details).
For each dataset, we select 100 data collections consisting of $n=4$ columns, by sampling three different variables $X_1, X_2, X_3$
uniformly at random without replacement among the set of all triplets then adding the output variable $Y$.
We report the mean and standard deviation of the attack accuracy averaged over 10 runs of 100 data collections each.

Contrary to the synthetic results above, the correlations in our real-world datasets are not uniform. 
Instead they are, to a different extent depending on the dataset, skewed towards zero. 
This means that a large fraction of the target correlations belong to the ``low'' bin $[-1/3, 1/3)$. 
Beyond this it also means that the correlation constraints  $\rho(X_1, Y)=\cos\theta_1$ and $\rho(X_2, Y)=\cos\theta_2$ are very small, thus leading to intervals $R=[\cos(\theta_1+\theta_2), \cos(\theta_1-\theta_2)]$ that are close to, but not equal to $[-1, 1]$ (e.g., $R=[-0.99, 0.99]$). 
These intervals thus encompass the ``low'' bin fully, and the other bins almost fully. 
For these two reasons, the model-less attack achieves on real-world datasets an ``artificially'' high accuracy.
One solution would be to balance the test distribution. However, we instead prefer to show results for $N_B=5$ bins. 
This indeed reduces the artificial advantage of the model-less attack while maintaining the generality of the evaluation. 

\begin{table}[!ht]
\centering
  \resizebox{\textwidth}{!}{  
\begin{tabular}{clcccc}
\toprule
\multirow{2}{*}{\textbf{Number of bins}} & \multirow{2}{*}{\textbf{Dataset}} & \textbf{Random} & \textbf{Model-less} & \multicolumn{2}{c}{\textbf{Model-based attack}} \\
& & \textbf{guess} & \textbf{attack} & Logistic Regression & MLP \\
\midrule
\multirow{3}{*}{$N_B=3$}& Fifa19 & 33.3 & 60.2 (5.0) & 91.2 (3.6) & 78.8 (4.5) \\
& C \& C & 33.3 & 73.6 (2.7)  & 86.0 (3.6) & 75.6 (4.0) \\
& Musk & 33.3 & 67.8 (5.6) & 82.0 (3.2) & 56.3 (6.2) \\
\midrule
\multirow{3}{*}{$N_B=5$}& Fifa19 & 20.0 & 29.4 (5.1) & 79.1 (3.6) & 61.2 (6.1) \\
& C \& C & 20.0 & 27.6 (3.2)  & 70.6 (3.8) & 56.0 (5.3) \\
& Musk & 20.0 & 28.7 (4.5) & 72.0 (5.5) & 41.7 (6.4) \\
\bottomrule
\end{tabular}
}
\caption{\textbf{Results of our correlation inference attacks on three real-world datasets.} We abbreviate ``Communities and Crime'' to ``C \& C'' for spacing reasons.}
\label{table:real_dataset_evaluation}
\end{table}

Table \ref{table:real_dataset_evaluation} shows that, despite its advantage, the model-less attack is outperformed by our model-based attack in five out of six cases. In Fifa19, it achieves an accuracy of 91.2\% against LR models and 77.9\% against MLP models, strongly outperforming the model-less attack which reaches 60.2\%. Our attack similarly outperforms the model-less attack in the Communities and Crime dataset, achieving and accuracy of 86.0\% against the LR models and 75.6\% against the MLP models compared to the 73.6\% reached by the model-less attack. Finally, while our attack achieves an accuracy of 82.0\% on LR models, it is outperformed by the model-less attack when it comes to MLP models in the Musk dataset (56.3\% vs 67.8\%).

However, as soon as we increase the number of bins, removing the artificial advantage of the model-less attack, our attack vastly and consistently outperforms the model-less attack. 
On Fifa19, it achieves an accuracy of 79.1\% against LR models and 61.2\% against MLP models, strongly outperforming the model-less attack which reaches 29.4\%. 
It similarly outperforms the model-less attack on the Communities and Crime and Musk datasets. 
On Communities and Crime it achieves an accuracy of 70.6\% against the LR models and 56.0\% against the MLP models compared to the 27.6\% reached by the model-less attack. 
On Musk, it achieves 72.0\% against the LR models and 41.7\% against the MLP models compared to the 28.7\% reached by the model-less attack.

\subsection{Attribute inference attack}
\label{subsec:results:attribute-inference-attack}
We have so far shown that dataset correlations, which are fundamental summary statistics about a dataset, can be extracted from a machine learning model trained on it.
We here argue that the ability of an attacker to accurately extract statistical information can enable other attacks, in particular allowing weaker attackers to infer membership or attributes from trained machine learning models.
More specifically, we present an attribute inference attack that uses as building blocks the correlations between input variables extracted from the model using our attack.

We assume the first variable in the dataset $X_1$ to be sensitive and consider an adversary that aims to recover the sensitive attribute value $x_1$ of a \textit{target record} $(x_1, x_2, \ldots, x_{n-1}, y) \in D_T$ using black-box access to a model $\mathcal{M}_T$ trained on $D_T$.
In line with the literature~\cite{fredrikson2014privacy,mehnaz2022your,jayaraman2022attribute}, we assume that the adversary knows the partial record $(x_2, \ldots, x_{n-1})$, the label $y$, and the one-way marginals of the attributes.

The intuition behind our attack is that the correlation between the sensitive variable $X_1$ and the other variables, when combined with knowledge of the partial record $(x_2, \ldots, x_{n-1}, y)$, can lead to accurate inference of the sensitive attribute $x_1$.
We propose an attack that extracts the correlations between the input variables  $\rho(X_i, X_j), 1 \leq i < j \leq n-1$ from the model using our model-based attack, under default scenario \textbf{S2}, and uses them to generate synthetic data satisfying these correlations.
We then search for (approximate) matches of the partial record $(x_2, \ldots, x_{n-1}, y)$ in the synthetic data and return as prediction $\hat{x_1}$ the average value of the sensitive attribute values of matched records.
We refer the reader to Sec.~\ref{subsec:materials-methods:attribute-inference} for the details.

We evaluate our correlation-inference-based attribute inference attack (CI-AIA) on the Fifa19 dataset~\cite{fifa192019}, randomly selecting 1000 data collections consisting of $n=4$ columns.
For each data collection, we train a Logistic Regression model and run the attack against 500 randomly selected records (see Sec.~\ref{appendix:attribute-inference-attack} in Supplementary Materials for the details).
To evaluate the attack performance, we divide the range of the sensitive attribute $X_1$ into three bins corresponding to the tertiles of the distribution.
Thus, a majority baseline attains an accuracy of roughly 33\%.

We compare our attack with five attacks from previous works: the first AIA of Fredrikson et al.~\cite{fredrikson2014privacy}, CSMIA by Mehnaz et al.~\cite{mehnaz2022your}, the threshold-based attack of Yeom et al.~\cite{yeom2018privacy}, and the CAI and WCAI methods of Jayaraman and Evans~\cite{jayaraman2022attribute}.
These methods are described in Supplementary Materials~\ref{appendix:attribute-inference-attack}.
We further compare our approach with two other baselines: a na\"ive approach that returns a random sample from the one-way marginal of the sensitive attribute $X_1$, which is known to the adversary, and another approach, referred to as $X_1 - Y$ \textbf{correlation}, that runs our CI-AIA approach on the variables $X_1$ and $Y$ alone.
Its goal is to quantify the leakage from the adversary knowledge of $\rho(X_1,Y)$ without access to the model.
Indeed, knowledge of the correlation $\rho(X_1,Y)$ can lead to better than random inference of the attribute $x_1$ since $y$ is known.
This baseline acts as a lower bound for CI-AIA, as the latter additionally assumes access to the model.

\begin{table}[!htbp]
    \centering
    \begin{tabular}{lc}
    \toprule
         \textbf{Method}& \textbf{Accuracy} \\
    \midrule
    (Ours) CI-AIA & \textbf{49.7 $\pm$ 1.0}  \\
    \midrule
    Fredrikson et al. \cite{fredrikson2014privacy} & 38.4 $\pm$ 0.8  \\
    CSMIA (Mehnaz et al.~\cite{mehnaz2022your}) &  46.5 $\pm$ 0.9 \\
    Yeom et al.~\cite{yeom2018privacy} & 38.5 $\pm$ 0.8\\
    CAI (Jayaraman and Evans~\cite{jayaraman2022attribute}) & 46.5 $\pm$ 0.9\\
    WCAI (Jayaraman and Evans~\cite{jayaraman2022attribute}) & 41.7 $\pm$ 0.8\\
    \midrule
    $X_1-Y$ correlation & 41.7 $\pm$ 0.8  \\
    Marginal prior & 35.3 $\pm$ 0.6 \\ 
    \bottomrule
    \end{tabular}
    \caption{\textbf{Comparison between CI-AIA and other attribute inference attacks on Fifa19.} We report the attack accuracy averaged over 1000 runs (with 95\% confidence interval).}
    \label{tab:table_aia}
\end{table}

Table~\ref{tab:table_aia} shows that our novel correlation inference-based attribute inference attack, referred to as CI-AIA, outperforms previous works and baselines, demonstrating how correlations can be used as building blocks for other attacks.
Our attack achieves an accuracy of 49.7\%, significantly better than CSMIA (Mehnaz et. al.~\cite{mehnaz2022your}), which only reaches 46.5\%, on par with CAI (Jayaraman and Evans~\cite{jayaraman2022attribute}), and the method of Fredrikson et al.~\cite{fredrikson2014privacy}, which reaches 38.4\% and is on par with Yeom et al.~\cite{yeom2018privacy}.

Our attack also outperforms the $X_1-Y$ correlation baseline, confirming our intuition that the additional correlations $\rho(X_1, X_i), i>1$ extracted from the model using our attack lead to more accurate attribute inference.

\section{Discussion}
\label{sec:discussion}

In this work, we study a new type of leakage in machine learning (ML) models, the leakage of correlations between input variables, proposing the first correlation inference attack against ML models.
Our results show that a model remembers more information than previously thought, raising privacy and confidentiality concerns whenever dataset correlations are sensitive information.
We develop two different attacks, \textit{model-less} and \textit{model-based}, that together allow to correctly quantify the leakage from the model.
We perform an extensive evaluation of our attacks, showing that models leak correlations of their training dataset with high accuracy, in both synthetic and real-world datasets.
We also show that mitigations such as limiting the number of queries that can be made to the model, their precision, or training the models with differential privacy guarantees do not help.
Finally, we show how correlations extracted using our attack can be used as building blocks for attribute inference attacks, in particular enabling weak adversaries to develop more powerful attacks than previously thought.

\textbf{Attack motivation.}
To further emphasize the privacy interest of correlation information, we here describe several use cases where correlation leakage might represent a privacy violation or could lead to individual harm. 
First, we consider the example presented in the introduction: a scoring model for depression which leaks that patients living in the inner city are more likely to have used illegal substances, an information that could be used to target them. 
Second, we consider a model controlling the signalization on the highway.
We assume that it was trained on data collected from vehicles in a specific area and that there is a positive correlation between the darkness of the car and the probability of accident in the area. 
If the model leaks this information to an insurance company operating in the area, this could lead to higher insurance premium for darker cars in the area. 
Third, we consider a model that leaks the correlation between age and income of people in its training dataset. Furthermore, assume that the model developer claims the model to be trained on representative data. Our correlation inference attack can be used as part of an auditing process, to surface a situation where although the training dataset is balanced along each individual attribute (e.g., age and income taken in isolation), it is not balanced on their intersection, e.g., \textit{in the training dataset} being older is excessively correlated with income.

\textbf{When does the leakage happen?} Our results (Fig.~\ref{fig:grid-attack} and Fig.~\ref{fig:impact-of-largest-constraint}) suggest that (1) models leak the correlation between input variables \textit{more} when these variables are highly correlated (positively or negatively) with the output variable $Y$ than when these variables are not correlated with $Y$, but that (2) leakage still occurs even when there is little to no correlation between these variables and the target variable.
The second finding might seem particularly surprising in the case of $n=3$ variables, where  the resulting model makes random predictions since all its input variables ($X_1$ and $X_2$) are statistically independent to the target variable $Y$ (uncorrelation implies statistical independence for Gaussian variables). The apparent contradiction can however be explained by the fact that a model making random predictions is not the same as the model not having any statistical dependency on its training dataset. 
Indeed, while the optimal model is only intended to learn $P(Y|X=x)$, current learning techniques such as stochastic gradient descent optimize the expected loss computed over records of the joint distribution $P(X, Y)$. Thus, the distribution of the input variables $P(X)$ influences the model. Whether the leakage we observe is a necessary condition for learning to happen and thus, whether robust defenses might exist, remains an open problem.

We now analyze some of the assumptions made on the adversary knowledge.

\textbf{Access to auxiliary data.} Our approach does not require access to the private dataset $D_T$, nor to a dataset drawn from a similar distribution.
Such detailed knowledge of the data is a common assumption in the literature on inference attacks and has led to strong results as the adversary uncertainty is reduced, e.g., ~\cite{ganju2018property,zhang2021leakage,carlini2022membership}.
We however believe that real-world adversaries are unlikely to have detailed knowledge of the data.
Our goal is thus to explicitly extract information about the data distribution and use it to infer attributes of individual records.
While in this work we have assumed a Gaussian copula prior on the underlying distribution (allowing us to control for correlations), future work could explore different priors to improve our results.

\begin{figure}[!ht]
\centerline{\includegraphics[width=\textwidth]{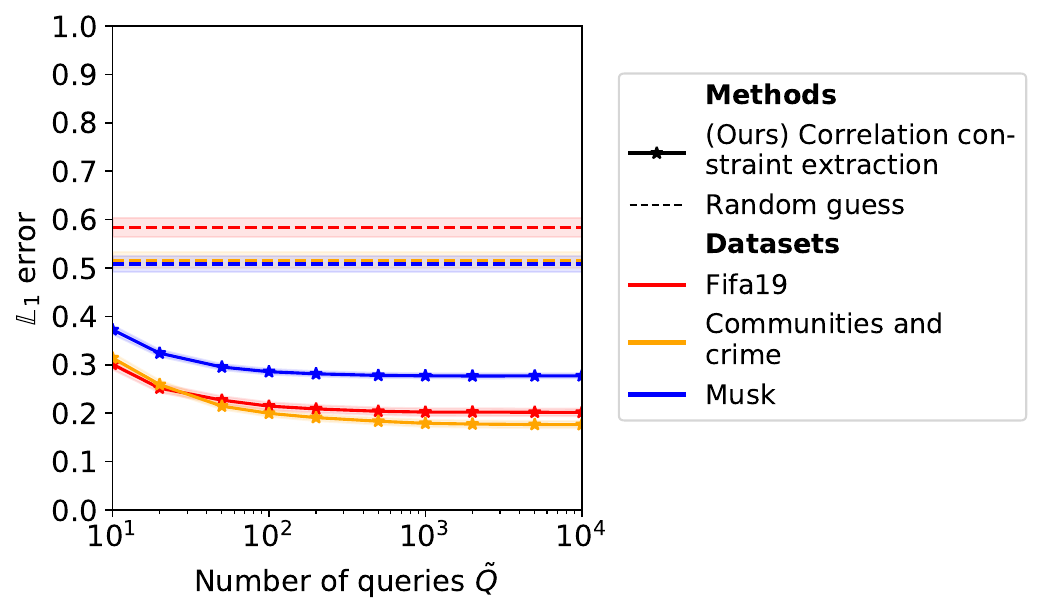}}
\caption{\textbf{Correlation constraint extraction from LR models for different number of queries $\tilde{Q}$.}
We compute the $\mathbb{L}_1$ error between the inferred and the ground-truth values of the correlations: $\frac{1}{n-1}\sum_{i=1}^{n-1} |\hat{\rho}(X_i, Y)-\rho(X_i, Y)|$ (mean with 95\% confidence interval). 
}
\label{fig:corr-extraction}
\end{figure} 

\textbf{Knowledge of the correlation constraints.} Our default scenario \textbf{S2} assumes the adversary to know $(\rho(X_i,Y))_{i = 1 ,...,n-1}$.
We show that these can be extracted from the model using a small number of queries $\tilde{Q}$.
More specifically, we sample $\tilde{Q}$ records $(x_1,...,x_{n-1})$ independently from the marginals $F_1,\ldots,F_{n-1}$, then query the model to retrieve the predicted label $y$.
We then estimate $(\hat{\rho}(X_i, Y))_{i=1,\ldots,n-1}$ on these records. 
We compute the $\mathbb{L}_1$ error of the predictions, defined as $\frac{1}{n-1}\sum_{i=1}^{n-1} |\hat{\rho}(X_i, Y)-\rho(X_i, Y)|$, over 500 data collections of $n=4$ columns, and compare it with the $\mathbb{L}_1$ error of a random guess in $[-1, 1]$.
Fig.~\ref{fig:corr-extraction} shows our approach to perform much better than the random guess, reducing the $\mathbb{L}_1$ error from 0.584 to 0.214 with as little as 100 queries to Logistic Regression models trained on the Fifa19 dataset.
The performance of our approach improves as the number of queries $\tilde{Q}$ increases, but stabilizes early at $\tilde{Q}=100$.

\textbf{Knowledge of precise marginals.} We have so far assumed the adversary to have precise knowledge of the marginals,  discretizing them for simplicity into $G=100$ sub-intervals of equal size.
Fig. \ref{fig:incr_nbr_marginal_bins} in the Supplementary Materials shows that decreasing $G$ and, by doing so, reducing the adversary's ability to sample from the marginals with fine granularity, has a very small impact on the performance of our attack.
The accuracy on Fifa19 decreases very slowly from 91.2\% ($G=100$) to 82.9\% ($G=5$), much higher than the model-less attack at 60.2\%.

\begin{figure}[!ht]
\centerline{\includegraphics[width=\textwidth]{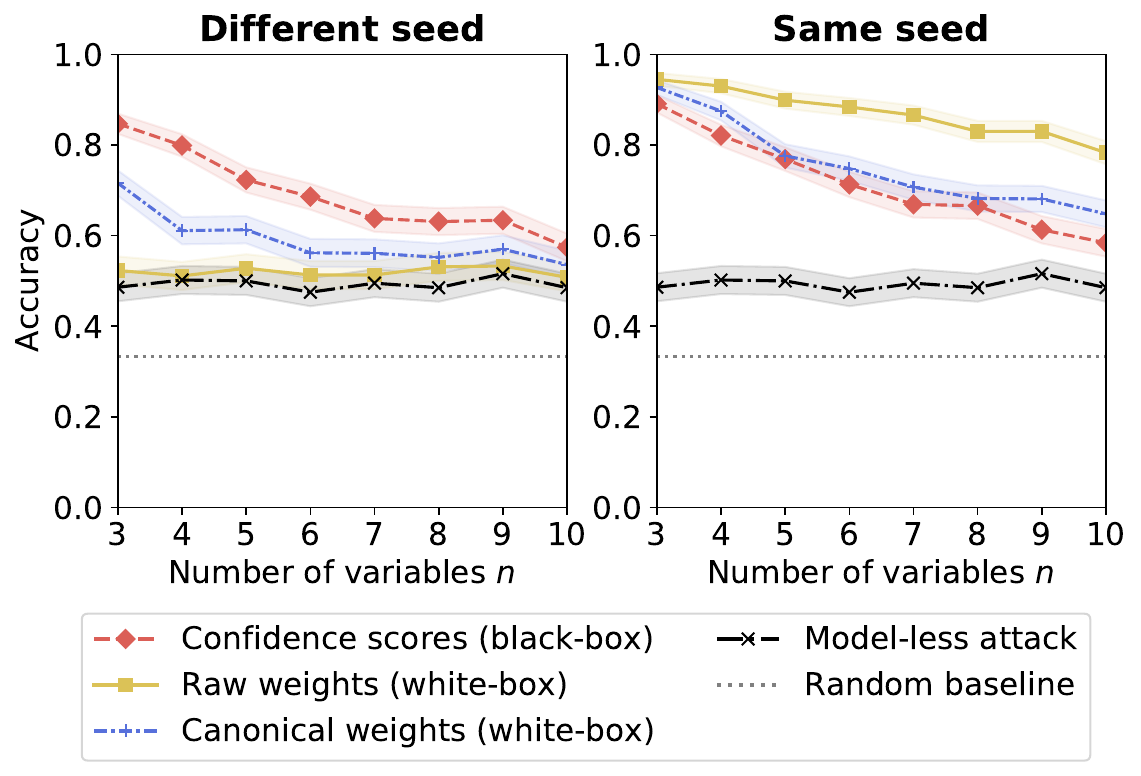}}
\caption{\textbf{Comparison between black-box and white-box CIAs against MLPs when the adversary uses different seeds (left) or the same seed (right) as the target model for training the models used to generate the meta-classifier's training dataset}. We report the mean attack accuracy (with 95\% confidence interval).}
\label{fig:impact-of-seed-mlp}
\end{figure} 

\textbf{Knowledge of the model seed.}
We believe it more realistic to assume that the adversary does not know the seed used to initialize the training algorithm of MLP models, including their weights.
This assumption leads to more uncertainty, as the models may end up in different local minima compared to the target model and may behave less similarly than expected.
This is not an issue for LR models, as they minimize a convex objective having a global optimum~\cite{balle2022reconstructing}.
Fig.~\ref{fig:impact-of-seed-mlp}A and B compare the performances of black-box and various white-box correlation inference attacks against MLP models, when changing, respectively keeping the seed constant between models used to generate the meta-classifier's training dataset and target models.
We refer the reader to the Supplementary Materials ~\ref{appendix:white_box_results} for implementation details of the white-box attacks.
While the black-box attack gains little from the seed knowledge (up to 4.6 p.p), the white-box attack reaches an astonishing 94.5\% for $n=3$ variables, 88.4 for $n=6$, and 78.3\% for $n=10$ variables, the last being 29.8 p.p. above the model-less attack.
It now also outperforms the attack on LR models for all $n\geq4$, confirming our hypothesis that more weights (e.g. 432 for the MLP vs 10 for the LR when $n=10$) can indeed encode more information.
Finally, the canonical sorting of weights, proposed by Ganju et al.~\cite{ganju2018property} to enhance the performance of the meta-classifier is, in this setting, somewhat effective when the seed changes.
However, its performance is still much lower than the black-box attack. 
It however achieves the opposite effect of destroying some of the information present in the raw weights when the seed is kept constant.

\section{Materials and methods}
In this section, we describe in detail our methodology for our correlation inference attack (Sec.~\ref{subsec:materials-methods:correlation-inference}) and our correlation inference-based attribute inference attack (Sec.~\ref{subsec:materials-methods:attribute-inference}).

\subsection{Correlation inference attack}
\label{subsec:materials-methods:correlation-inference}

We describe in detail the various building blocks of our methodology for the model-less and model-based attacks.
More specifically, we first describe our algorithms for generating correlation matrices under constraints (Sec.~\ref{subsec:materials-methods:generate-correlation-matrices}).
Second, we describe how we generate synthetic dataset conditionally on a correlation matrix for arbitrary one-way marginals (Sec.~\ref{subsec:materials-methods:sampling-dataset}).
Third, we describe the details of our model-based attack (Sec.~\ref{subsec:materials-methods:model-based-attack}).

\subsubsection{Generating correlation matrices under constraints}\label{subsec:materials-methods:generate-correlation-matrices}

We adapt the algorithm of Numpacharoen and Atsawarungruangkit~\cite{numpacharoen2012generating}, which we include in the Supplementary Materials~\ref{appendix:sample_correlation_matrix} (Alg. S1), to our problem of generating matrices under constraints given by the adversary's knowledge.
We notice that the coefficients of the first column of $B$ ($\cos~\theta_{i, 1}$ for  $i=2, \ldots, n-1$) are identical to those of $C$ and are free parameters. 
We start by reordering the variables to $Y, X_1, \ldots, X_{n-1}$ to generate $\rho(X_1, X_2)$ first (and explain below why). 
Under \textbf{S2}, we set the coefficients of the first column of $B$ (and $C$) equal to the correlation constraints $(\rho(X_i, Y))_{i=1, \ldots, n-1}$.
This amounts to replacing line 6 from Alg. S1 (highlighted in red) with $c_{i,1}\leftarrow \rho(X_{i-1}, Y)$.
Under \textbf{S1}, we will set the first two coefficients equal to the correlation constraints $\rho(X_1, Y)$ and $\rho(X_2, Y)$, initializing the others uniformly at random within $[-1, 1]$. 
We replace line 6 from Alg. S1 correspondingly.

\begin{algorithm}[!ht]
\caption{\textsc{SampleCorrMatrix-AttackS2}}
\label{alg:sample_correlation_matrix_s2}
    \begin{algorithmic}[1]
        \Inputs{$n$: Number of variables.\\
        $constraints=(\rho(X_i, Y))_{i=1,\ldots,n-1}$: Correlation constraints imposed by the adversary's knowledge.}
        \Output{$C \in \mathbb{R}^{n\times n}$ : A valid correlation matrix satisfying $c_{i,n}=\rho(X_i, Y)$, for $i=1, \ldots, n-1$. }
        \State{\commblack{Randomly shuffle the constraints.}}
        \State{$\sigma \gets [1, 2] + random\_permutation([3, \ldots, n-1])$}
        \State{$constraints \gets \sigma(constraints)$}
        \State{$C \gets \textsc{SampleCorrMatrix}(n,  constraints)$}
        \State{\commblack{Reorder the variables as $X_1, \ldots, X_{n-1}, Y$.}}   
        \State{$\sigma' \gets [2,  \ldots, n] + [1]$}
        \State{$C \gets reorder\_columns(reorder\_rows(C, \sigma'), \sigma')$}
        \State{\commblack{Revert the shuffling of constraints.}}
        \State{$\sigma_n \gets \sigma^{-1} + [n]$}
        \State{$C \gets reorder\_columns(reorder\_rows(C, \sigma_n), \sigma_n)$}
    \end{algorithmic}
\end{algorithm}

We then shuffle the variables 3 to $n-1$ by applying a random permutation. 
Indeed, na\"ively applying Alg. S1 would generate the correlations using the default variable ordering (in our case, $Y, X_1, \ldots, X_{n-1}$). 
While the correlation generated first, $\rho(X_1, X_2)$, is uniformly distributed within its boundaries, the correlation generated last, $\rho(X_{n-2}, X_{n-1})$, is not, because the previously sampled correlations restrict its range. 
More specifically, its conditional distribution follows a bell-shaped curve.
Shuffling the variables reduces this bias~\cite{numpacharoen2012generating}.
We choose to keep the target pair $X_1$ and $X_2$ in the first position, to guarantee a uniform prior over the distribution of the target correlation $\rho(X_1, X_2)$.
Indeed, we assume the adversary does not know how $\rho(X_1, X_2)$ is distributed within its boundaries. 
Furthermore, we do not want to bias the values towards the center of the interval, as would be the case if the value was not generated first.
Alg.~\ref{alg:sample_correlation_matrix_s2} details how we carefully sequence the shuffling, the column reordering, and the inversion of the shuffling for scenario \textbf{S2} (see Alg. S2 in the Supplementary Materials \ref{appendix:sample_correlation_matrix_s1} for scenario \textbf{S1}).

Alg.~\ref{alg:sample_correlation_matrix_s3} details our procedure to sample the unknown value $\rho(X_1, X_2)$ for scenario \textbf{S3}, when all the other correlations are known.
We start by placing the target pair $(X_1, X_2)$ in the last position. 
Denote by $C'$ the partial correlation matrix after reordering the variables (filling the unknown $c'_{n, n-1}$ with a dummy value).
We sample the last coefficient $c'_{n, n-1}$ uniformly within its bounds, which we determine by reverse-engineering part of the spherical parametrization $B$.
More specifically, we compute all the low triangular coefficients except for $b_{n, n-1}$ and $b_{n, n}$, which cannot be determined absent knowledge of $c'_{n, n-1}$.
This allows us to compute the bounds of $c'_{n, n-1}$ and sample it uniformly within them.

We discuss in the Supplementary Materials~\ref{appendix:sample_correlation_matrix} an alternative, optimization-based approach to generate correlation matrices under constraints, that we tried in preliminary experiments and we found to be ineffective.

\begin{algorithm}[htbp!]
\caption{\textsc{SampleCorrMatrix-AttackS3}}
\label{alg:sample_correlation_matrix_s3}
    \begin{algorithmic}[1]
        \Inputs{$n$: Number of variables.\\
        $C$: Adversary knowledge having $c_{1,2}$ and $c_{2, 1}$ filled with a dummy value.}
        \Output{$D \in \mathbb{R}^{n\times n}$ : A valid correlation matrix satisfying $d_{i, j}=c_{i, j}$, $\forall (i, j) \in \{ 1, \ldots, n\}^2 \setminus \{ (1, 2), (2, 1)\}$.}
        \State{\commblack{Reorder the variables as $Y, X_{n-1}, \ldots, X_1$.}}
        \State{$\sigma \gets [n, \ldots, 1]$}
        \State{$C' \gets reorder\_columns(reorder\_rows(C, \sigma), \sigma)$}
        \State{$B' B'^T \gets \textbf{CholeskiDecomposition}(C'_{1:n-1, 1:n-1})$ \commblack{$B' \in \mathbb{R}^{(n-1)\times(n-1)}$}}
        \State{$B \gets 0$; $B_{1:n-1, 1:n-1} \gets B'$ \commblack{$B \in \mathbb{R}^{n \times n}$}}
        \State{$b_{n, 1}\gets c'_{n, 1}$ \commblack{Start filling the last row of $B$.}}
        \For{$i \in \{2,\cdots,n-2\}$}
        \State{$b_{n,i} \gets b'_{n, i} - B_{i, 1:i}(B_{n, 1:i})^T  / b_{i, i}$}
        \EndFor
        \State{$b_{n, n-1} \gets \sqrt{1 - \sum_{j=1}^{n-2} b_{n, j}^2}$}
        \State{\commblack{Compute the boundaries $m \pm l$ of $c'_{n, n-1}$.}}
        \State{$m, l \gets B_{n-1, 1:n-1} (B_{n, 1:n-1})^T, b_{n-1, n-1} b_{n, n-1}$}
        \State{\commblack{Sample uniformly within the bounds.}}
        \State{$c'_{n, n-1} = c'_{n-1, n} \gets \mathcal{U}(m-l, m+l)$}
        \State{\commblack{Reorder the variables as $X_1, \ldots, X_{n-1}, Y$.}}
        \State{$D \gets reorder\_columns(reorder\_rows(C', \sigma))$}
    \end{algorithmic}
\end{algorithm}

\subsubsection{Generating synthetic data given a correlation matrix for arbitrary one-way marginals}
\label{subsec:materials-methods:sampling-dataset}
Alg.~\ref{alg:shift_correlation_constraints} details our  heuristic to modify the correlation constraints.
More specifically, it determines $V'$ such that, on average, synthetic datasets $D$ generated from matrices matching $V'$ minimize the difference $\max_{(i, j) \in P} |E[C(D)_{i,j}] - (c_T)_{i, j}|$ between their empirical correlations $C(D)$ and the correlation constraints.
For conciseness, we focus on the default scenario $\textbf{S2}$, where $V=(\rho(X_i, Y))_{i=1, \ldots, n-1}$ (of which scenario $\textbf{S1}$ is a particular case).

\subsubsection{Model-based attack}
\label{subsec:materials-methods:model-based-attack}

Our model-based attack exploits access to a machine learning model $\mathcal{M}_T$ trained on the target dataset $D_T$ in order to extract the target correlation. 

First, we generate $K$ correlation matrices $C^1, \ldots, C^K$ satisfying $V$ when the marginals are standard normals (and $V'$ for arbitrary marginals).
In each scenario, the correlation matrices are sampled using the corresponding algorithm. 
For each correlation matrix $C^k$, we generate a synthetic dataset $D^k$ whose correlations approximately match $V$ and whose marginals match $F_i, i=1, \ldots, n$ as described in Sec.~\ref{subsec:materials-methods:sampling-dataset}.

\begin{algorithm}[!htbp]
\caption{\textsc{ModifyCorrelationConstraints}}
\label{alg:shift_correlation_constraints}
    \begin{algorithmic}[1]
        \Inputs{
        $n$: Number of variables. \\
        $V = (\rho(X_i,Y))_{i=1,\ldots,n-1} \in \mathbb{R}^{n-1}$: Correlation constraints. \\
        $(F_i)_{i=1, \ldots, n}$: One-way marginals. \\
        $S$: Number of synthetic datasets used to estimate the gap. \\
        $N_D$: Size of the synthetic datasets. \\
        $e$: Error tolerance. \\
        $M$: Maximum number of iterations.   
        }
        
        \Output{$V' \in \mathbb{R}^{n-1}$: Modified correlation constraints. 
        }
        \Initialize{$V' \gets V$}
        \For{$m = 1, \ldots, M$}
            \State{$\bar{V} \gets zeros(n-1)$}
            \For{$s = 1, \ldots, S$}
            \State{$C \gets \textsc{SampleCorrMatrix-AttackS2}(n, V')$}
            \State{\commblack{Synthetic dataset whose correlations approximately match $V'$.}}
            \State{$D^s \gets [\textsc{SampleFromGaussianCopulas}($ \\
            $n, C,F_1, \ldots, F_n)$ for $d = 1, \ldots, N_D]$}
            \State{$C^s \gets correlation\_matrix(D^s)$}
            \State{$\bar{V}_i \gets \bar{V}_i + \frac{C^s_{i,n}}{S}$ for $i=1,\ldots,n-1$}
            \EndFor
        \State{$gap \gets \max_{i=1,\ldots,n-1}|\bar{V}_i - V_i|$}
        \If{$gap < e$}
            \State{\textbf{break}}
        \Else
            \State{\commblack{Update the  constraints.}}
            \State{$V'_i \gets max(-1, min(1, V'_i + \frac{V_i - \bar{V}_i}{2} )$ for $i=1,\ldots,n-1$}
        \EndIf
        \EndFor
    \end{algorithmic}
\end{algorithm}

Second, we train a model $\mathcal{M}^k$ on dataset $D^k$, for every $k=1, \ldots, K$.
The models are trained using the same algorithm and hyperparameters as the target model, but a different seed.
We present in Sec.~\ref{sec:discussion} results of our attack when the adversary knows the seed.

Third, we generate a synthetic \textit{query dataset} $D_{\text{query}}$ in the same way as above, and use it to extract black-box features from the models.
More specifically, after the $K$ models have been trained, we pass the records of $D_{\text{query}}$ through the model $\mathcal{M}^k$ for $k=1,\ldots, K$, and retrieve the confidence scores for the first $L-1$ classes (the last one is redundant with the others as they all sum to 1).
Denoting by $Q$ the size of $D_{\text{query}}$, we thus extract a total of $Q\times(L-1)$ features from each model.
In this paper, we assume for simplicity that the models are binary classifiers ($L=2$), meaning that we extract $Q$ features from each model.
We concatenate them to form the \textit{output feature vectors} $O_k, l=1, \ldots, K$.
Our hypothesis is that these features are likely to encode information about the dataset correlations, e.g. that a model makes more confident predictions on records resembling its own training dataset.

Fourth, given a dataset $D$, we denote by $C(D)$ the correlation matrix computed on the dataset.
We label each feature vector $O_k$ with the correlation $L_k = C(D^k)_{1, 2}$ of the corresponding synthetic dataset, discretized over $B$ classes.
We then train a meta-classifier $\mathcal{A}: \mathbb{R}^Q \rightarrow \{ 1, \ldots, B\}$ on $D_{\text{meta}} = \{ (O_k, L_k), k=1,\ldots,K \}$ to infer the correlation given features extracted from the model.

Finally, we deploy the meta-classifier on the target model $\mathcal{M}_T$. 
The adversary's guess is the output class $\mathcal{A}(O_T)$ predicted by the meta-classifier on the feature vector $O_T$, consisting of the outputs of $\mathcal{M}_T$ on the query dataset $D_{\text{query}}$.

\subsection{Attribute inference attack}
\label{subsec:materials-methods:attribute-inference}
We describe our attribute inference attack aiming to infer the sensitive attribute value $x_1$ of a target record $(x_1, \ldots, x_{n-1},y) \in D_T$ using black-box access to a model $\mathcal{M}_T$ trained on a private dataset $D_T$.
Our method consists of three steps:

\underline{Step 1:} We execute our correlation inference attack independently against each target pair $(X_i, X_j), 1 \leq i < j \leq n-1$ under default scenario \textbf{S2}. 
The output of the attack is a correlation bin $b_{i, j} \in \{1, \ldots, N_B\}$.
To evaluate our attack on real-world datasets having arbitrary marginals, we run the attack using the \textsc{ModifyCorrelationConstraints} heuristic.
We denote by $V'_{i, j} \in \mathbb{R}^{n-1}$  the modified correlation constraints returned by the heuristic.

\underline{Step 2:} We sample synthetic datasets whose empirical correlations belong to the inferred bins, while also  matching the correlation constraints $(\rho(X_i, Y))_{i=1, \ldots, n-1}$.
Sampling synthetic data under constraints given by the adversary knowledge requires to modify the correlation constraints $\rho(X_i, Y),i=1,\ldots,n-1$, as described in Sec.~\ref{subsec:materials-methods:sampling-dataset}.
We here use  the \textit{average} of modified constraints used previously to attack each of the pairs: $V' = \frac{2}{(n-1)(n-2)} \sum_{1 \leq i < j \leq n-1} V'_{i, j}$.
We generate $S'$ synthetic datasets under these modified constraints, then select among them the datasets $D$ whose empirical correlations $C(D)_{i, j}$ belong to the inferred bin $b_{i, j}$, for every $1 \leq i < j \leq n-1$, discarding the rest.
We concatenate the datasets together into a larger dataset $D_{\text{synth}}$.

\underline{Step 3:} We select all the values of the sensitive attribute $x_1^1, \ldots, x_1^H$ belonging to records that approximately match the partial record $(x_2, \ldots, x_{n-1}, y)$.
We search for approximate matches as follows. First, we discretize the marginals of non-sensitive variables $X_2, \ldots, X_{n-1}$ into $G$ sub-intervals of same size and denote by $b_2, \ldots, b_{n-1}$ the size of the sub-intervals.
Second, we retrieve all the records $(x_1', \ldots,x_{n-1}', y')$ in $D_{\text{synth}}$ such that: (1) $|x_i - x'_i| \leq m_i b_i, i=2,\ldots,n-1$ and (2) $y = y'$, where $m_i$ denotes the resolution of the search around the partial record's attribute $x_i$.
For instance, for $m_2=1$ and $b_2=0.2$ we  only select records having $x'_2 \in [x_2-0.2, x_2+0.2]$.
If no record is found, we increase the resolution parameters by increments $\delta_i$: $m_i \gets m_i + \delta_i, i=2,\ldots,n-1$, until at least one record is found. 
We return the average sensitive attribute of matched records $\hat{x}_1 = \frac{1}{H} (\sum_{h=1}^H x_1^h)$ as our prediction. We also explore ablations of our attack using different formulas for predicting $x_1$ given the partial record $(x_2,\cdots,x_{n-1},y)$ and the synthetic data generated under Step 2, including variations of Jayaraman and Evans's~\cite{jayaraman2022attribute} formula, none of which significantly improve the attack's performance. Summary of the results can be found in Appendix \ref{appendix:attribute-inference-attack}.

As for the $X_1-Y$ \textbf{correlation} baseline, we run steps 2 and 3 of our CI-AIA approach, described above, on variables $X_1$ and $Y$, but not step 1 as there is no need to extract additional correlations from the model. 

\section{Related work}
\label{sec:related-work}
\label{sec:related_work}

\textbf{Information leakage by ML models} has, so far, been mostly studied  through the lens of \textit{individual records}. This includes membership inference attacks~\cite{shokri2017membership,salem2018ml,yeom2018privacy,nasr2019comprehensive,choquette2021label,carlini2022membership} aiming to infer if a record was used to train a model -- and to a lesser extent, attribute inference~\cite{fredrikson2014privacy,fredrikson2015model,mehnaz2022your} and reconstruction~\cite{salem2020updates,balle2022reconstructing} attacks -- aiming to reconstruct respectively, partially and entirely, a record. 
While potentially very invasive from a privacy perspective, attacks against individual-level records commonly rely on powerful adversaries, with partial~\cite{nasr2019comprehensive} or near-complete~\cite{balle2022reconstructing} knowledge of the dataset, or with access to a dataset sampled from the same distribution. 
We here focus on an adversary aiming to infer correlations without access to real records, which requires rethinking the current design of inference attacks.
In particular, we develop a model-less attack as a strong baseline and use Gaussian copulas to sample synthetic tabular datasets that we use to generate the meta-classifier's training dataset.

\textbf{Property inference attacks (PIA)} study the leakage of macro properties of the dataset. 
Ateniese et al.~\cite{ateniese2015hacking} introduced property inference attacks (PIAs), albeit by another name, and applied them to Hidden Markov Models and Support Vector Machines.
They show, for instance, that it is possible to distinguish the dialect on which a speech recognition model was trained. 
To this end, they train a meta-classifier on \textit{white-box} features extracted from models trained in the same way as the target model, a methodology now known as \textit{shadow modeling}~\cite{shokri2017membership}.
Ganju et al.~\cite{ganju2018property} showed that extending this attack to multilayer perceptrons (MLP) is more challenging due to their permutation equivalence property and propose to use weight-based neuron sorting and set-based meta-classifiers; we use the former in our white-box experiments on MLPs. 
Zhang et al.~\cite{zhang2021leakage} proposed the first \textit{black-box} property inference attack, making use of the model's output probabilities computed on a dataset of real records from the same distribution.
Finally, Suri and Evans~\cite{suri2022formalizing} proposed a formal definition of distribution inference attacks that unifies the property inference attacks (PIA) of previous works, which focused on the leakage of the proportion of samples along a certain attribute.
The authors extend the definition to other properties such as the average node degree of a graph.

Our work differs substantially from previous work in PIAs. 
First, we focus, by design, on explicit micro characteristics of the dataset $X$ rather than on external macro properties of the record (e.g., the gender label of an image).  
Second, we demonstrate how dataset-level information, here correlations, can be used as building blocks for more accurate attribute inference attacks.
Third, our attack, contrary to PIAs, does not require access to a subset of the dataset or, at least, a dataset sampled from a similar distribution. This would indeed mean that the adversary could readily compute the correlations or, at least, have a very good prior on them. 
We here do not assume access to private data for approximating the distribution, developing a methodology for inferring correlations which relies on synthetic data instead of private data.
More specifically, the models we use to generate the training dataset of the meta-classifier are trained exclusively on synthetic records.
This makes the attack accessible to weaker adversaries than those of previous works.
Fourth, previous works mainly focus on a categorical choice between \textit{very distinct} properties. 
For instance, Ganju et al.~\cite{ganju2018property} aim to infer whether the proportion of older faces is equal to (precisely) 23\% or 37\% or if exactly 0\% or 87\% of the faces where from white people~\cite{ganju2018property}. 
By contrast, we frame the correlation inference task as a prediction task \textit{covering the entire range of possible values}.
While Zhou et al.~\cite{zhou2021property} aim to address this shortcoming by framing property inference as a regression task, they only apply their attack to Generative Adversarial Networks which would, by design, reflect the statistical properties of the training dataset making a correlation attack trivial. 

\textbf{Attribute inference attacks (AIA)} aim to recover the sensitive attribute of a target record. Friedrikson et al.~\cite{fredrikson2014privacy} proposed the first AIA against ML models. 
They applied their attack to linear regression models available as a black-box and aimed to infer a patient's genotype. 
Subsequent work~\cite{fredrikson2015model} extended this attack to make use of confidence scores and applied it to decision trees.
Yeom et al.~\cite{yeom2018privacy} explored the connection between membership and attribute inference and proposed a black-box attack relying on a membership inference oracle. 
Mehnaz et al.~\cite{mehnaz2022your} then proposed new black-box attacks: (1) a confidence-based attack dropping the prior work's assumption that the adversary knows the one-way marginals of the dataset and (2) a label-based attack assuming the adversary has access to a large fraction of the training dataset (excluding the sensitive attribute). 

In contemporaneous work, Jayaraman and Evans~\cite{jayaraman2022attribute} propose two extensions of the black-box AIA of Yeom et al.~\cite{yeom2018privacy} and a white-box AIA.
Our black-box correlation inference-based AIA (CI-AIA) is fundamentally different from their black-box AIAs, because we study a weaker adversary who does not know the data distribution via access to auxiliary data.
Instead, we (1) extract information about the distribution from the model (the correlation between the input variables), (2) use this information to generate synthetic data, and (3) predict the sensitive attribute using the synthetic data.
Table~\ref{tab:table_aia} shows our CI-AIA to outperform their black-box AIAs when compared fairly in our weak adversary setting.
We further note that our step (3) applied to synthetic data bears similarity with Jayaraman and Evans's~\cite{jayaraman2022attribute} prediction rule applied to auxiliary data.
To understand if their prediction rule can further improve our CI-AIA performance, we substituted their prediction rule to step (3) of our attack while leaving our steps (1) and (2) unchanged.
We show in Appendix~\ref{appendix:attribute-inference-attack} that this does not lead to better performance than our CI-AIA.

\textbf{Synthetic data-based shadow modeling} is an under-explored research area. 
Shokri et al.~\cite{shokri2017membership} were the first to use synthetic data to train shadow models. 
They developed a hill-climbing algorithm to generate synthetic records. Their approach is based on the intuition that a model would be more confident on records that are similar to the training dataset. 
Salem et al.~\cite{salem2018ml} argued that this method is only efficient for datasets of binary records and explored the use of shadow datasets from different sources.
This is however not a suitable option when aiming to infer summary statistics of a dataset as it requires shadow datasets to span the entire range of possible values for the statistics.

\section*{Acknowledgments}
Ana-Maria Cre\c{t}u did most of her work while she was a PhD candidate at Imperial College London. The authors would like to thank Florimond Houssiau for his feedback on the paper. We acknowledge computational resources and support
provided by the Imperial College Research Computing Service \url{http://doi.org/10.14469/hpc/2232}.

\subsection*{Funding}
Ana-Maria Cre\c{t}u and Yves-Alexandre de Montjoye were supported by the PETRAS National Centre of Excellence for IOT Systems Cybersecurity, funded by the UK EPSRC under grant number EP/S035362/1.

\subsection*{Author contributions}
A.-M.C. and F.G. designed and implemented the attacks, designed and implemented the experiments, generated the figures, and wrote the article. Y.-A.d.M. designed the attack and experiments and wrote the article.

\subsection*{Competing interests}
All authors declare no competing interests.

\subsection*{Data and Materials Availability}
The three real-world datasets we report results on are publicly available: Fifa19~\cite{fifa192019}, Communities and Crime~\cite{communitiesandcrime2011}, and Musk~\cite{muskv21998}.
All other experiments are performed on synthetic datasets.
Our source code allowing to generate the synthetic datasets and reproduce the results of our experiments is available at \url{https://doi.org/10.14469/hpc/14302} (permanent location) and   \url{https://github.com/computationalprivacy/ml-correlation-inference} (Github repository).

\bibliographystyle{plain}
\bibliography{bibliography}

\newpage
\appendix
\renewcommand{\thepage}{S\arabic{page}}
\renewcommand{\thesection}{S\arabic{section}}
\renewcommand{\thefigure}{S\arabic{figure}}
\renewcommand{\thealgorithm}{S\arabic{algorithm}}
\renewcommand{\thetable}{S\arabic{table}}
\setcounter{page}{1}
\setcounter{algorithm}{0}
\setcounter{figure}{0}
\setcounter{table}{0}
\begin{center}
\textbf{\Large Supplementary Materials for  ``Correlation inference attacks against machine learning models''}
\end{center}

\label{sec:appendix}
\startcontents[sections]
\printcontents[sections]{l}{1}{\setcounter{tocdepth}{2}}

\section{Details of the model-less attack}
\label{appendix:details_model_less_attack}
We present the complete details of the model-less attack, for preciseness and reproducibility. 
First, we define the $N_B$ classification bins  as the following intervals: 
\begin{gather} 
\begin{aligned}\Bigg[ \frac{2(b-1)-N_B}{N_B}, \frac{2b-N_B}{N_B} \Bigg) & \text{, for } b = 1, \ldots, N_B-1 \\
\Bigg[\frac{2(b-1)-N_B}{N_B}, \frac{2b-N_B}{N_B} \Bigg] & \text{, for } b = N_B
\end{aligned}
\end{gather}

Denoting by $[m_1, m_2]$ the range of possible values for target correlation $\rho(X_1, X_2)$, our model-less attack predicts the majority bin over this interval, distinguishing three cases:
\begin{enumerate}
\itemsep0em
    \item[\textbf{(C1)}] The interval is fully included inside a bin, i.e., $(2(b-1)-N_B)/N_B \leq m_1 \leq m_2 < (2b-N_B)/N_B$ for some $b=1,\ldots, N_B$. We predict the bin $b$.
    \item[\textbf{(C2)}] The interval $[m_1, m_2]$ partially covers two bins, but none entirely, i.e., $m_1 < (2b-N_B)/N_B \leq m_2 < (2(b+1)-N_B)/N_B$, for some $b=1, \ldots, N_B-1$. 
    We predict the bin $b$ if it has higher coverage than $b+1$ (i.e., if $(2b-N_B)/N_B - m_1 > m_2 - (2(b+1)-N_B)/N_B$), and $b+1$ otherwise.
    \item[\textbf{(C3)}] At least one bin is fully covered by the interval $[m_1, m_2]$. We predict one of the bins that we sample uniformly at random.
\end{enumerate}

\section{Analysis of regions}
\label{appendix:shape_analysis}

We analyze in detail the sets of constraints $(\rho(X_1, Y), \rho(X_2, Y))$ satisfying that there is only one possible bin for $\rho(X_1, X_2)$.
We consider $0 \leq \theta_1, \theta_2 \leq \pi$ such that $\rho(X_1, Y)=\cos\theta_1$ and $\rho(X_2, Y)=\cos\theta_2$.
As a reminder, the $N_B=3$ classification bins are $[-1, -1/3)$,  $[-1/3, 1/3)$, and $[1/3, 1]$ and the interval attainable by $\rho(X_1, X_2)$ is $[\cos (\theta_1 + \theta_2), \cos(\theta_1 - \theta_2)]$. 

\begin{itemize}
    
    \item \textbf{Negative bin: $\cos (\theta_1 - \theta_2) \leq -\frac{1}{3}$}. It follows that $\theta_1 - \theta_2 \geq \arccos (-\frac{1}{3})$ if $\theta_1 \geq \theta_2$ and $\theta_2 - \theta_1 \geq \arccos (-\frac{1}{3})$ otherwise.
    
    \item \textbf{Low bin: $-\frac{1}{3} \leq \cos (\theta_1 + \theta_2) \leq \cos (\theta_1 - \theta_2) < \frac{1}{3}$}.
    We distinguish 4 subcases: 
    
    \textbf{(I)} $\theta_1 \geq \theta_2$ and $0 \leq \theta_1 + \theta_2 \leq \pi$. 
    It follows that $\theta_1 + \theta_2 \leq \arccos \big(-\frac{1}{3}\big)$ and $\theta_1 - \theta_2 > \arccos \big(\frac{1}{3}\big)$.
    
    \textbf{(II)}  $\theta_1 < \theta_2$ and $0 \leq \theta_1 + \theta_2 \leq \pi$. 
    It follows that $\theta_1 + \theta_2 \leq \arccos \big(-\frac{1}{3}\big)$ and $\theta_2 - \theta_1 >\arccos \big(\frac{1}{3}\big)$.
    
    \textbf{(III)} $\theta_1 \geq \theta_2$ and $\pi \leq \theta_1 + \theta_2 \leq 2\pi$. It follows that $\theta_1 + \theta_2 \geq \pi + \arccos \big(\frac{1}{3} \big)$ and $\theta_1 - \theta_2 > \arccos \big(\frac{1}{3} \big)$.
    
    \textbf{(IV)} $\theta_1 < \theta_2$ and $\pi \leq \theta_1 + \theta_2 \leq 2\pi$. It follows that $\theta_1 + \theta_2 \geq \pi + \arccos \big(\frac{1}{3} \big)$ and $\theta_2 - \theta_1 > \arccos \big(\frac{1}{3} \big)$.
    
     \item \textbf{Positive bin: $\frac{1}{3} < \cos (\theta_1 + \theta_2)$} It follows that  $\theta_1 + \theta_2 < \arccos (\frac{1}{3})$ if $\theta_1 + \theta_2 \leq \pi$ and to $\theta_1 + \theta_2 > \pi + \arccos (-\frac{1}{3})$ if $\pi \leq \theta_1 + \theta_2 \leq 2\pi$.
\end{itemize}

\section{Algorithm to sample a valid correlation matrix}
\label{appendix:sample_correlation_matrix}
Alg.~\ref{alg:sample_correlation_matrix} details an implementation of the algorithm to sample a valid correlation matrix by Numpacharoen and Atsawarungruangkit~\cite{numpacharoen2012generating}.
We refer the reader to the original work for a slightly modified version which correctly handles the numerically unstable cases when the range of attainable values for $c_{i,j}$ is very small.

\begin{algorithm}[!ht]
\caption{\textsc{SampleCorrMatrix}~\cite{numpacharoen2012generating}. We highlight in red the statement which we modify when calling this algorithm inside Alg. 1 of the main paper and Alg. S2. The modified statement for each case is provided in Sec. 4.1.1 and is omitted here for brevity. Code comments are written in blue.}
\label{alg:sample_correlation_matrix}
    \begin{algorithmic}[1]
        \Inputs{$n$: Number of variables.
        }
        \Output{$C \in \mathbb{R}^{n\times n}$: A valid correlation matrix.
        }
        \Initialize{
        $C \gets 0 \;;\; B \gets 0$}
        \State{\comm{Randomly initialize the first column.}}
        \For{$i \in \{2, \ldots, n\}$}
            \State{\textcolor{red}{$c_{i,1} \gets \mathcal{U}(-1, 1)$ \comm{$\cos\theta_{i,1}$}}}
            \State{$b_{i,1} \gets c_{i,1}$}
            \State{$b_{i,j} \gets \sqrt{1-c_{i,1}^2}$ \; \text{for} \; $j \in \{2, \ldots, i\}$ \comm{$\sin\theta_{i,1}$}}
        \EndFor
        
        \For{$i \in \{2,\cdots,n\}$}
        \For{$j \in \{1,\cdots,i\}$}

        \State{\comm{Compute the bounds $m_{i,j} \pm l_{i,j}$.}}
        \State{$m_{i,j} \gets B_{i,1:j-1} (B_{j, 1:j-1})^T$}
        \State{$l_{i,j} \gets b_{i,j} b_{j, j}$}

        \State{\comm{Sample uniformly within bounds.}}
        \State{$c_{i,j} \gets \mathcal{U}(m_{i,j}-l_{i,j}, m_{i,j} + l_{i,j})$}
        
        \State{\comm{Update $B$.}}
        \State{$aux \gets \frac{c_{i,j}-m_{i,j}}{l_{i,j}}$\comm{$\cos~\theta_{i,j}$.}}
        
        \State{$b_{i,j} \gets b_{i,j} * aux$}

        \State{$b_{i,k} \gets \sqrt{1-aux^2} \; \text{for} \; k \in \{j+1, \ldots, n\}$}
        \EndFor
        \EndFor
    \State{$C \gets C + C^T + I_n$}
    \end{algorithmic}
\end{algorithm}

\textbf{Alternative approaches.} The spherical parametrization of correlation matrices provides a principled and effective approach for sampling valid correlation matrices. 
In contrast, the characterization given by properties \textbf{P1-P4} does not easily translate into an effective algorithm for constructing correlation matrices. 
In preliminary experiments, we framed the correlation matrix generation problem as a constrained optimization program. We initialized a $n\times n$ matrix by setting each element uniformly at random between -1 and 1, then projected it into the space of positive semi-definite matrices having all diagonal entries equal to 1 and all elements between -1 and 1. We used the DCCP library~\cite{shen2016disciplined}.
Due to numerical instability, the resulting matrices often did not satisfy property  \textbf{P4}, as they had at least one negative eigenvalue (40\% of the time for $n=4$ variables and decreasing steadily with $n$).
As a result, several instances of the optimization program are required to generate one valid correlation matrix, and their number increases with the number of variables $n$. In constrast, Alg.~\ref{alg:sample_correlation_matrix_s1} succeeds 100\% of the time, regardless of the number of variables. 
Furthermore, our approach is 10 times faster than one instance of the optimization program.
We also found that the correlations generated by the optimization program are less likely to cover the entire range of values that are theoretically attainable.

\section{Algorithm to sample a correlation matrix under scenario S1}
\label{appendix:sample_correlation_matrix_s1}

Alg. \ref{alg:sample_correlation_matrix_s1} is the implementation of the algorithm to sample a correlation matrix using the default scenario S1. It uses Alg. \ref{alg:sample_correlation_matrix} while carefully applying the random permutation according to the constraints from the S1 scenario.  

\begin{algorithm}[!ht]
\caption{\textsc{SampleCorrMatrix-AttackS1}. Code comments are written in blue.}
\label{alg:sample_correlation_matrix_s1}
    \begin{algorithmic}[1]
        \Inputs{$n$: Number of variables.\\
        $constraints=(\rho(X_1, Y), \rho(X_2, Y))$: Correlation constraints imposed by the adversary's knowledge.}
        \Output{$C \in \mathbb{R}^{n\times n}$ : A valid correlation matrix satisfying $c_{1,n}=\rho(X_1, Y)$ and $c_{2, n} = \rho(X_2, Y)$. }
        \State{$C \gets \textsc{SampleCorrMatrix}(n,  constraints)$}
        \State{\comm{Reorder variables as $X_1, X_2, X_{\sigma(3)}, \ldots, X_{\sigma(n-1)}, Y$.}}
        \State{$\sigma \gets [2, 3] + random\_permutation([4,  \ldots, n]) + [1]$}
        \State{$C \gets reorder\_columns(reorder\_rows(C, \sigma),\sigma)$}
    \end{algorithmic}
\end{algorithm}

\section{Algorithm to sample from Gaussian copulas}
\label{appendix:sample_from_gaussian_copulas}
Alg.~\ref{alg:sample_from_gaussian_copulas} details a procedure to obtain a sample from a $n$-dimensional Gaussian copulas distribution parametrized by a covariance matrix $\Sigma$ and one-way marginals $F_1, \ldots, F_n$.

\begin{algorithm}[!ht]
\caption{\textsc{SampleFromGaussianCopulas.} Code comments are written in blue.}
\label{alg:sample_from_gaussian_copulas}
    \begin{algorithmic}[1]
        \Inputs{$n$: Number of variables.\\
        $\Sigma \in \mathbb{S}_+^d$: A valid covariance matrix.\\
        $F_1, \ldots, F_n$: One-way marginals.
        }
        
        \Output{A sample $z=(z_1, \ldots, z_n)$ from a Gaussian copula distribution parametrized by $F_1, \ldots, F_n$ and $\Sigma$.}
        
        \State{ $A \gets \textbf{CholeskyDecomposition}(C)$ \comm{$C = A^{t}\cdot A$}}
        
        \State{$Z \sim \mathcal{N}(0,I_n)$}
        \State{$X \gets A^T Z$}
        \For{$i = 1, \ldots, n$}
        \State{$z_i \gets F^{-1}_i(X_i)$}
        \EndFor
        \State{$z \gets (z_1, \ldots, z_n)$}
    \end{algorithmic}
\end{algorithm}

\section{Experimental setup}
\label{appendix:experimental-setup}
\subsection{Target models}
\label{appendix:experimental-setup:target-models}
The target models we study in this paper are the Logistic Regression (LR) and the Multilayer Perceptron (MLP). 

For the LR models, we use the implementation provided by the Scikit-learn 0.24.1~\cite{pedregosa2011scikit} library with default parameters.
For the MLP models, we implement in Pytorch 1.10.0~\cite{paszke2019pytorch} an architecture consisting of two hidden layers of sizes 20 and 10 with ReLU nonlinearity. 
In the experiments on synthetic data (Sec.~\ref{subsec:results:grid-evaluation}-~\ref{subsec:results:mitigations}), we train the models using gradient descent. In the experiments on real-world datasets (Sec.~\ref{subsec:results:real-world-datasets}), we instead train the models using mini-batch gradient descent with a batch size of 128, as this leads to better target model accuracy.
The models are trained on 90\% of the samples for up to 100 epochs, stopping the models after 5 epochs of non-improving accuracy on the remaining 10\%. 
We use a learning rate of $\eta=0.05$.

\subsection{Datasets}
\label{appendix:experimental-setup:datasets}
We use three datasets in our real-world evaluation (Sec.~\ref{subsec:results:real-world-datasets}).

\textbf{Fifa19} \cite{fifa192019} contains 18207 records of 89 attributes each, describing physical attributes and performance statistics of football players. 
We discard the categorical attributes, keeping the continuous and discrete ordinal variables.
We select the \textit{Value} attribute, representing the amount of money a club payed for a player, as the output variable, discarding the other attributes predictive of it that would make classification trivial (e.g., \textit{Wage}).
We then remove the records with missing values, leaving us with 15917 records.
We binarize the output variable by mapping it to $Y=1$ if larger than the median value ($> 0.74$M) and to $Y=0$ otherwise.
Finally, we remove the duplicate attributes, which leaves us with 53 input attributes.

\textbf{Communities and Crime} \cite{communitiesandcrime2011} contains 2215 records of 147 attributes each, describing socio-economic, law enforcement, and crime statistics on communities in the US.  
We select the number of murders as the output variable of our machine learning models, discarding the other columns predictive of it (i.e., relating explicitly to crime). 
As input attributes, we select the 101 continuous and discrete ordinal attributes from the dataset that do not have missing values, discarding the rest.
We binarize the number of murders by mapping it to $Y=1$ if is at least 1, and to $Y=0$ otherwise. 

\textbf{Musk (v2)} \cite{muskv21998} contains 6598 records of 166 attributes each, describing different conformations of 165 molecules, together with a human expert-assigned label of ``musk'' or ``non-musk''. 
We train the models to classify conformations as ``musk'' ($Y=1$) or ``non-musk'' ($Y=0$). 
The dataset being heavily imbalanced with respect to the two classes, we balance it for simplicity,  leaving us with 2034 records. 

\subsection{Attack parameters}
\label{appendix:experimental-setup:attack-parameters}
Unless otherwise specified, throughout the paper we will consider the default attacker \textbf{S2}.
Given a target model $\mathcal{M}_T$, we train a meta-classifier on outputs extracted from $k=|D_{\text{meta}}|$ models.
We generate $K=5000$ synthetic datasets for target models $\mathcal{M}_T$ trained on $n=3, 4$ and $5$ variables and $K=10000$ synthetic datasets for $n=6$ to $10$ variables. 
In the experiments on synthetic data, the models are all trained on 1000 samples and the target model test accuracy is computed on 500 unseen samples.
In the experiments on real-world datasets, the target models are trained on all the samples available ($|D_T|=2215$ for Communities and crime and $|D_T|=2034$ for Musk), except for Fifa19, where we sample $|D_T|=2000$ records uniformly without replacement to keep the dataset sizes similar. 
The synthetic datasets are always generated to have a number of samples equal to the target dataset size $|D_T|$. 
We run the \textsc{ModifyCorrelationConstraints} heuristic (Alg.~\ref{alg:shift_correlation_constraints}) using parameter values $S=100$ and $N_D=|D_T|$, for a maximum of $M=10$ iterations and with an error tolerance of $e=0.01$.
We set the size of the query dataset $D_{\text{query}}$ equal to $Q=100$, and discuss in Sec~\ref{subsec:results:mitigations} the impact of this parameter.

We use a Logistic Regression as the meta-classifier for LR models, as it leads to similar accuracy as the MLP but takes less time to train.
We use an MLP as the meta-clasifier for MLP models, trained the same as stated in Sec.~\ref{appendix:experimental-setup:target-models} except that we use a batch size of 128, a learning rate of $\eta= 0.001$, $\mathbb{L}_2$ weight decay of 0.01, and early stopping after 10 epochs of non-improving accuracy on 10\% of held-out samples. 

\subsection{Model-less attack performance: number of samples}
\label{appendix:model-less-number-of-samples}
Throughout the experiments, for a fair comparison between the model-less and model-based attacks, we have used the same number of samples for the model-less and model-based attack, $K$.
We have opted for relatively large values of $K$, in the order of thousands (see Sec.~\ref{appendix:experimental-setup:attack-parameters} for the values used).
This is because the model-based attack, being machine learning-based, needs more samples to converge. We here examine how the performance of the model-less attack varies with the number of samples $K$.
We compute the accuracy of the model-less attack over $T=1000$ correlation matrices of $n=3$ variables using $K\in\{1, 2, 5, \ldots, 100, 200, 500, 1000\}$.

\begin{figure}[h]
\centering
\includegraphics[width=0.6\textwidth]{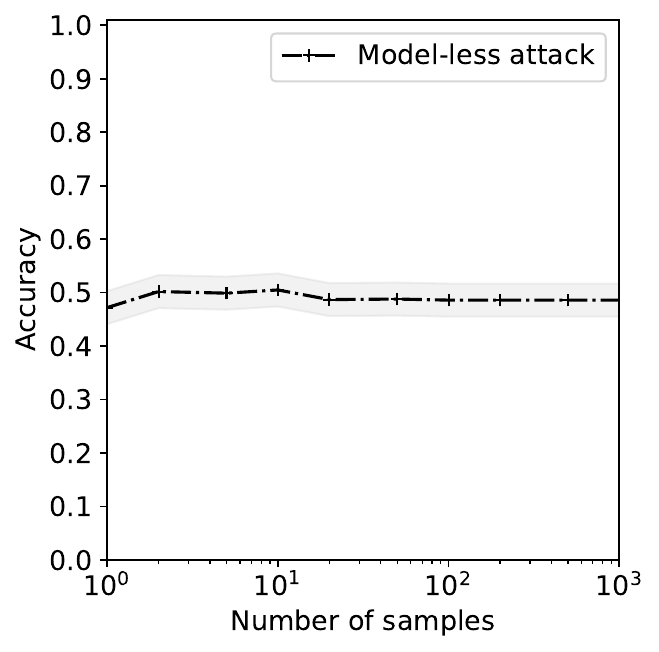}
\caption{\textbf{Impact of the number of samples $K$ on the model-less attack performance.} We report the accuracy (with 95\% confidence interval) over 1000 target models.}
\label{fig:model-less-analysis-samples}
\end{figure}

Fig.~\ref{fig:model-less-analysis-samples} shows that the accuracy of the model-less attack converges very quickly as a function of the number of samples. This is because the model-less attack predicts which of the $N_B=3$ bins is most covered by the segment $[\min(samples), \max(samples)]$, as described in Sec.~\ref{subsec:results:attack-method} - Model-less attack and Appendix \ref{appendix:details_model_less_attack} and this prediction is very robust to the number of samples used.

\section{White-box results}
\label{appendix:white_box_results}

We evaluate the performance of white-box correlation inference attacks using the model weights as features. 
The weights of the Logistic Regression are its coefficients (including the bias).
For the MLP, we explore two options: (1) using the \textit{raw weights} (coefficients of the linear layers, including the bias), flattened and concatenated into a single vector and (2) the \textit{canonical weights}, extracted in the same way from the model after sorting the neurons in each layer according to the sum of weights~\cite{ganju2018property}.

\begin{figure}[htp]
\centerline{\includegraphics[width=\textwidth]{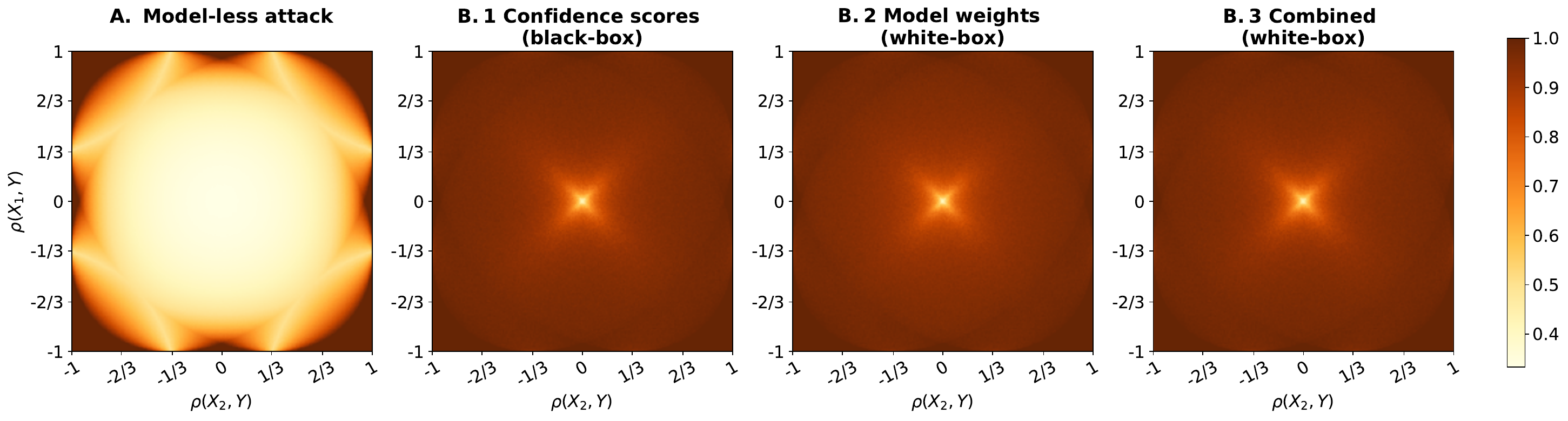}}
\caption{\textbf{Logistic regression: Comparison between black-box and white-box correlation inference attacks.} 
We report results for our model-less attack (A), our black-box model-based attack using as features the confidence scores (B. 1), and our white-box model-based attacks using as features the model weights (B. 2) or the combined weights and confidence scores (B. 3).
}
\label{fig:grid-attack-logreg}
\end{figure}

Fig.~\ref{fig:grid-attack-logreg} shows that the weights of Logistic Regression models yield similar performance to the confidence scores (95.1\% vs 95.6\%).
This is likely due to the small number of weights (three in total), whose information we believe to be redundant with the one contained in the confidence scores extracted from $Q=100$ records.

\begin{figure}[htp]
\centerline{\includegraphics[width=\textwidth]{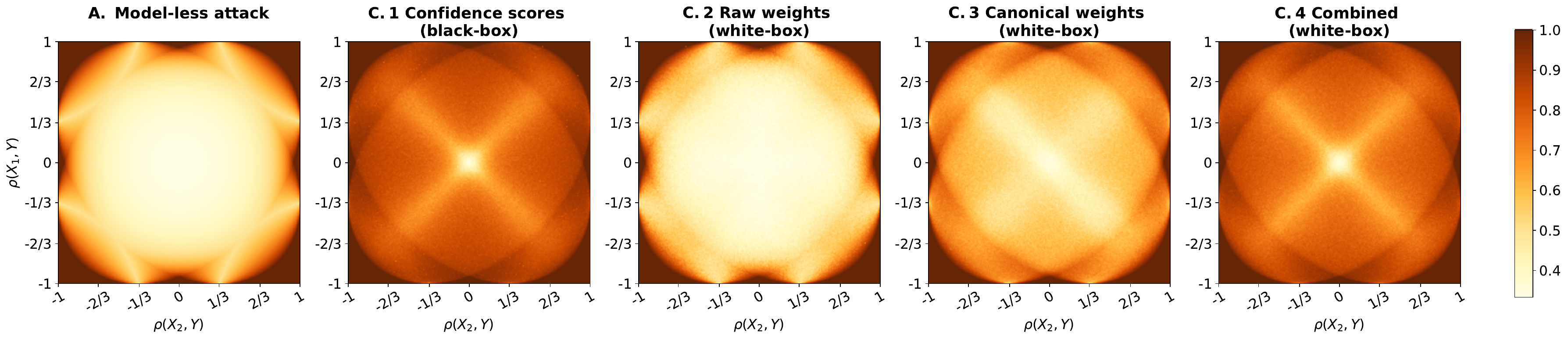}}
\caption{\textbf{Multilayer perceptron (MLP): Comparison between black-box and white-box correlation inference attacks.} 
We report results for our model-less attack (A), our black-box model-based attack using as features the confidence scores (C. 1), and our white-box model-less attacks using as features the raw model weights (C. 2), the canonical weights (i.e., the raw weights mapped to a canonical representation~\cite{ganju2018property}, C. 3), or the combined canonical weights and confidence scores (C. 4).
}
\label{fig:grid-attack-mlp}
\end{figure}

Fig.~\ref{fig:grid-attack-mlp} shows that the raw weights of MLP models do not leak more information than the model-less attack (55.0\%).
This negative result is in line with previous findings from the property inference literature~\cite{ganju2018property,zhang2021leakage} and can be attributed to the permutation equivalence property of MLPs~\cite{ganju2018property}, whereby neurons of internal layers can be permuted without changing the underlying function.
As a result, a meta-classifier operating on inputs in their default ordering (with $k!$ equivalent orderings being possible in each layer, where $k$ is the number of neurons in the layer) needs to learn a more difficult task.
We show that the canonical weights yield much better performance (65.5\%), albeit lower than the confidence scores. 

On both models, combining the model weights and the confidence scores (Fig.\ref{fig:grid-attack-logreg}-B.3 and Fig.~\ref{fig:grid-attack-mlp}-C.4) does not yield better results than using the confidence scores alone.

We refer the reader to Sec.~\ref{sec:discussion} for results of our white-box correlation inference attacks on MLP models when the attacker is assumed to know the seed used to initialize the target model's training algorithm.

\section{Additional results on synthetic datasets}

\begin{figure*}[!htbp]
\centerline{\includegraphics[width=\textwidth]{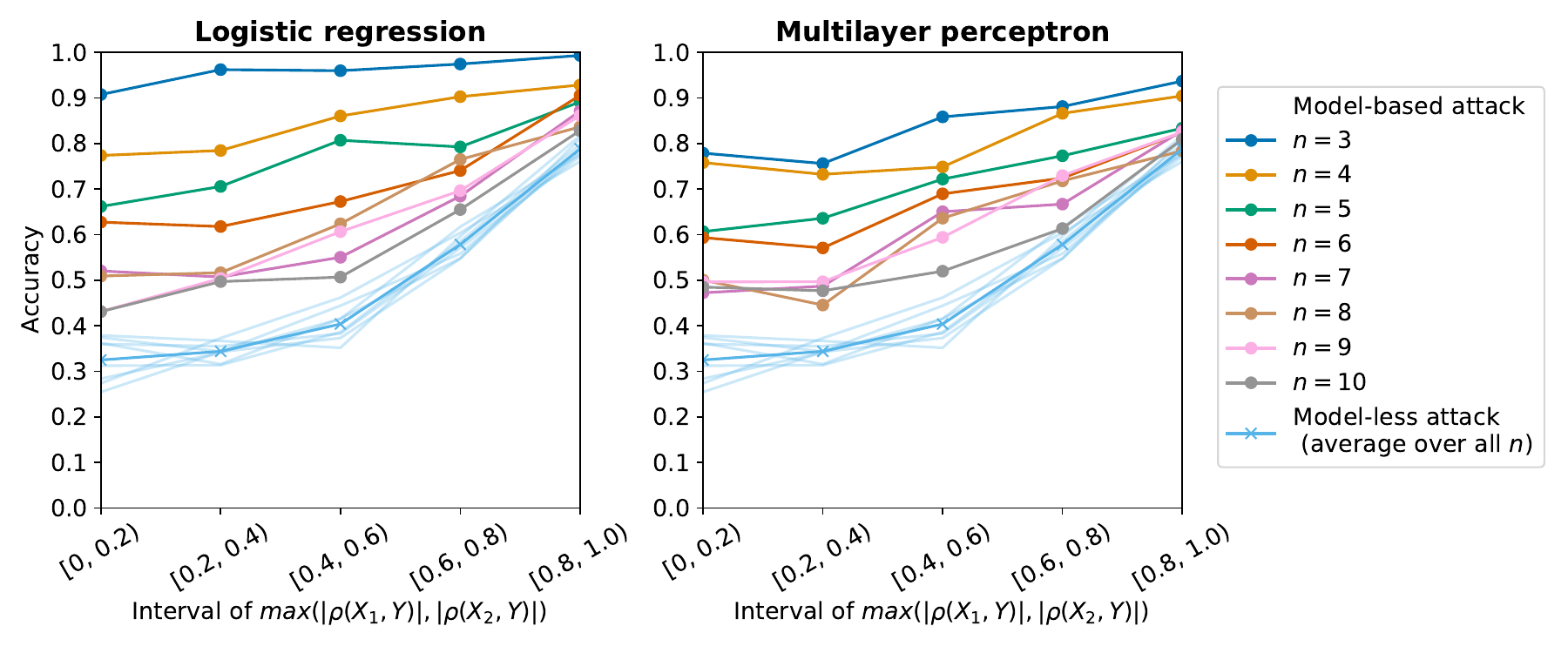}}
\caption{\textbf{Impact of the largest constraint  $\max(|\rho(X_1,Y)|,|\rho(X_2,Y)|)$ on the attack accuracy for different number of variables $n$.} We compare the results of our model-based and model-less attacks on logistic regression (left) and MLP models (right).}
\label{fig:impact-of-largest-constraint}
\end{figure*}

\begin{figure*}[!htbp]
\centerline{\includegraphics[width=\textwidth]{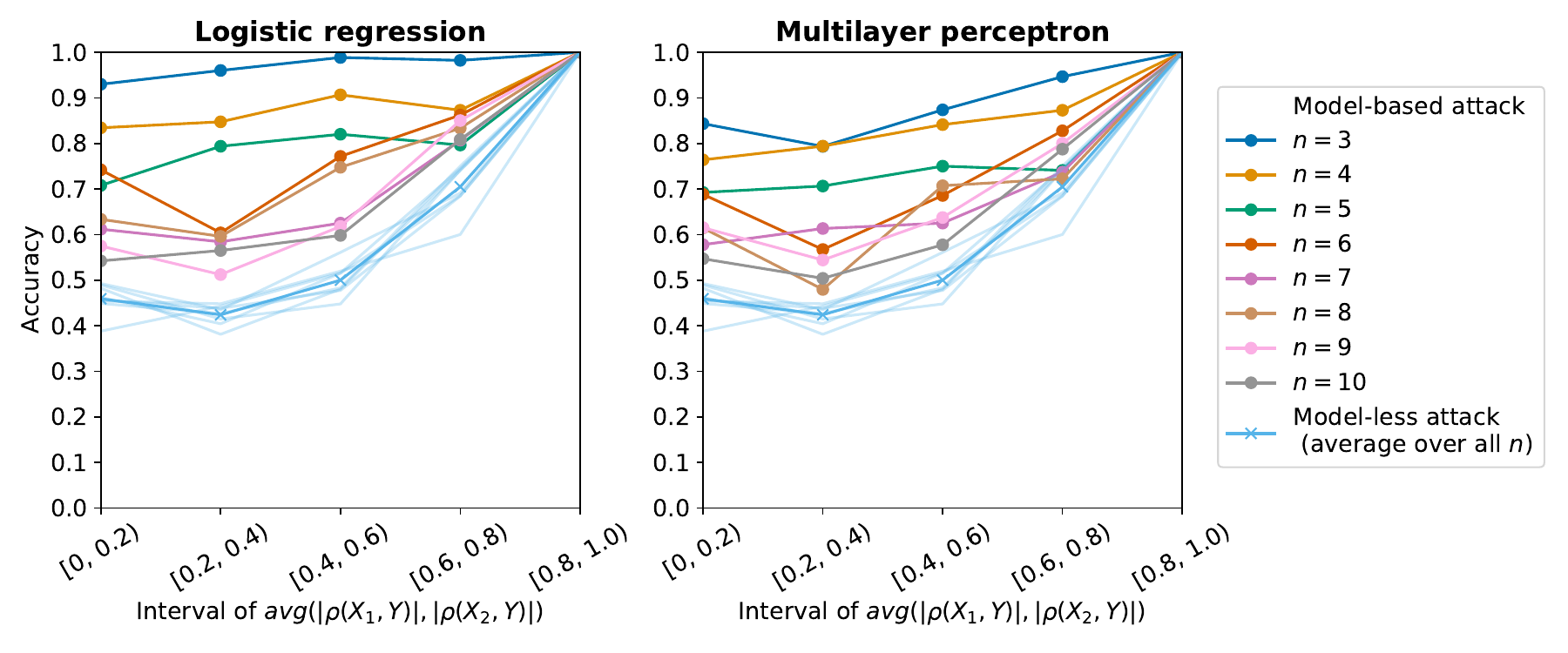}}
\caption{\textbf{Impact of the average constraint  $(|\rho(X_1,Y)|+|\rho(X_2,Y)|)/2$ on the attack accuracy for different number of variables $n$.} We compare the results of our model-based and model-less attacks on logistic regression (left) and MLP models (right).}
\label{fig:impact-of-average-constraint}
\end{figure*}

\begin{figure*}[!htbp]
\centerline{\includegraphics[width=\textwidth]{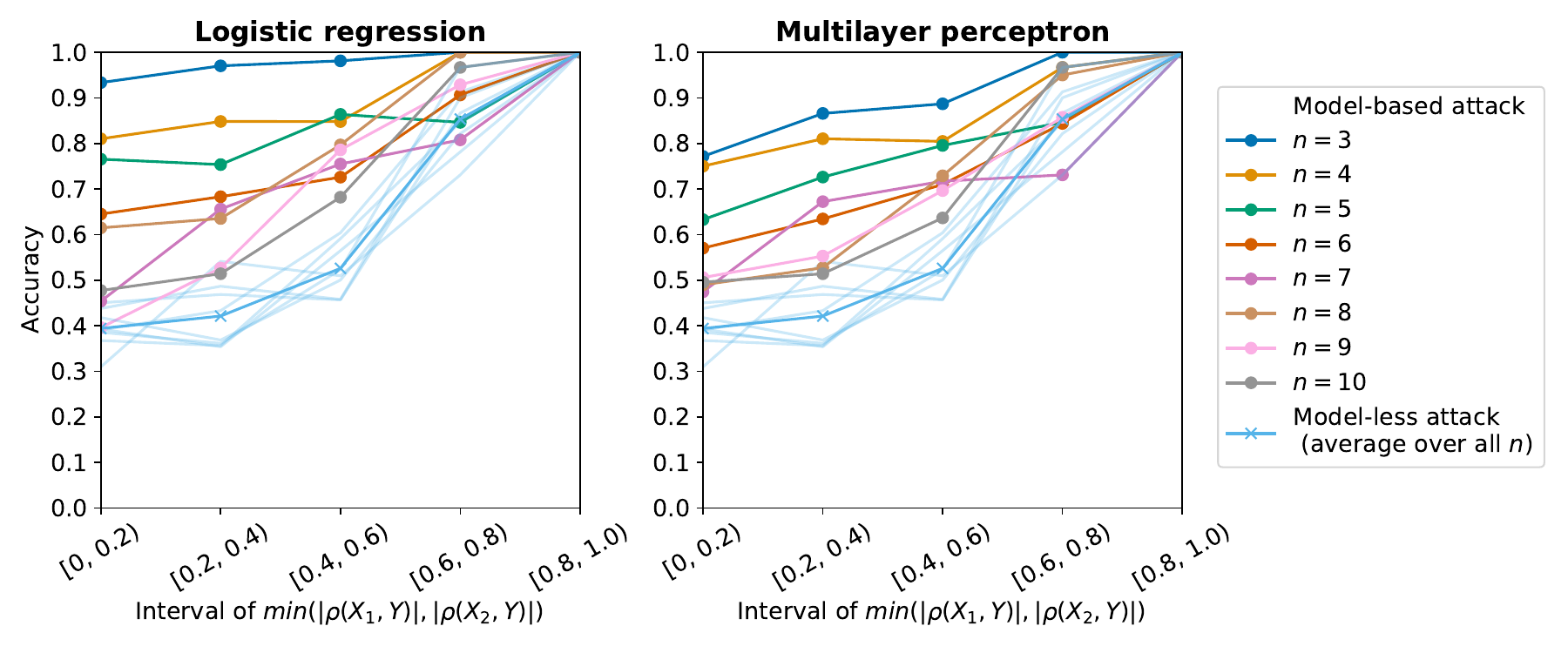}}
\caption{\textbf{Impact of the smallest constraint  $\min(|\rho(X_1,Y)|,|\rho(X_2,Y)|)$ on the attack accuracy for different number of variables $n$.} We compare the results of our model-based and model-less attacks on logistic regression (left) and MLP models (right).}
\label{fig:impact-of-smallest-constraint}
\end{figure*}

Fig. \ref{fig:impact-of-largest-constraint} shows the impact of the largest constraint $\max(|\rho(X_1, Y), \rho(X_2, Y)|)$ on the accuracy of our model-based and model-less attack.
For completeness, we also include results on the impact of the average constraint $(|\rho(X_1, Y), \rho(X_2, Y)|)/2$ (Fig.~\ref{fig:impact-of-average-constraint}) and of the smallest constraint $\min(|\rho(X_1, Y), \rho(X_2, Y)|)$ (Fig.~\ref{fig:impact-of-smallest-constraint}).

\begin{figure}[h]
\centerline{\includegraphics[width=0.6\textwidth]{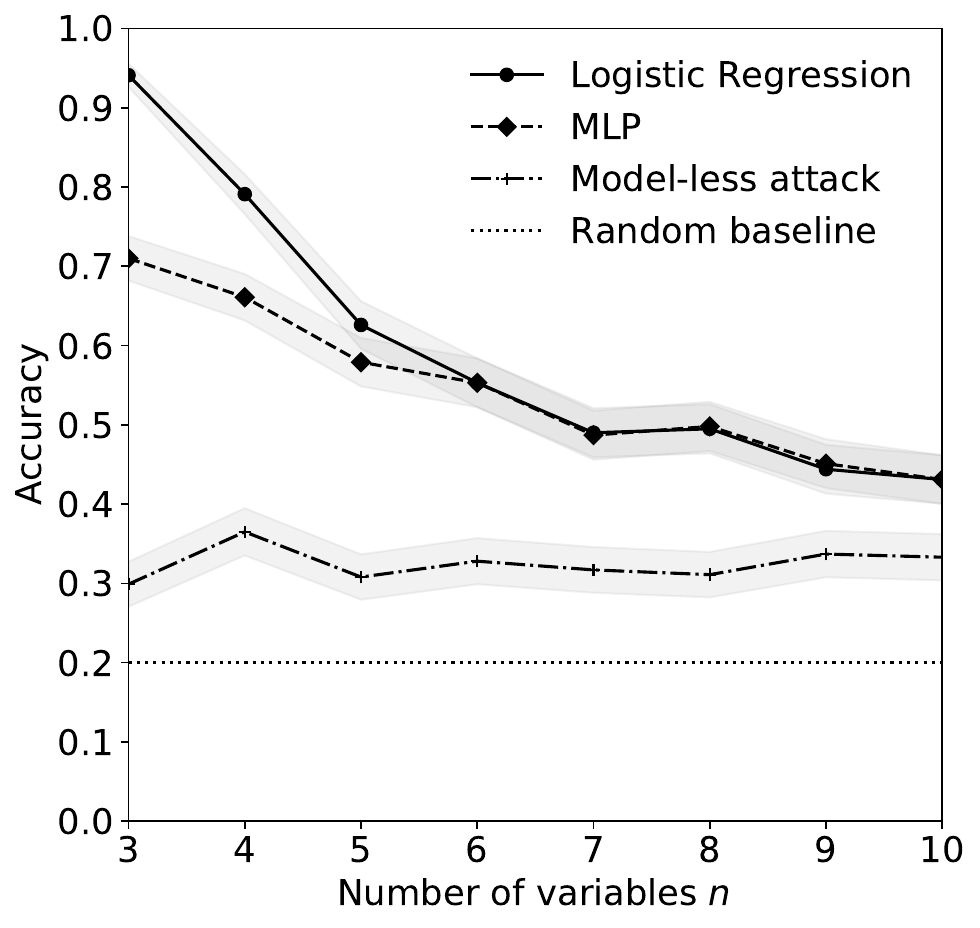}}
\caption{\textbf{Attack accuracy under scenario \textbf{S2} and $N_B=5$ correlation bins for different number of variables in the dataset $n$}. We report the accuracy (with 95\% confidence interval) over 1000 target models.}
\label{fig:rand-tar-attack-5-bins}
\end{figure}

Fig.~\ref{fig:rand-tar-attack-5-bins} shows results of our correlation inference attack using $N_B=5$ bins.

\begin{figure}[h]
    \centering
    \includegraphics[width=\textwidth]{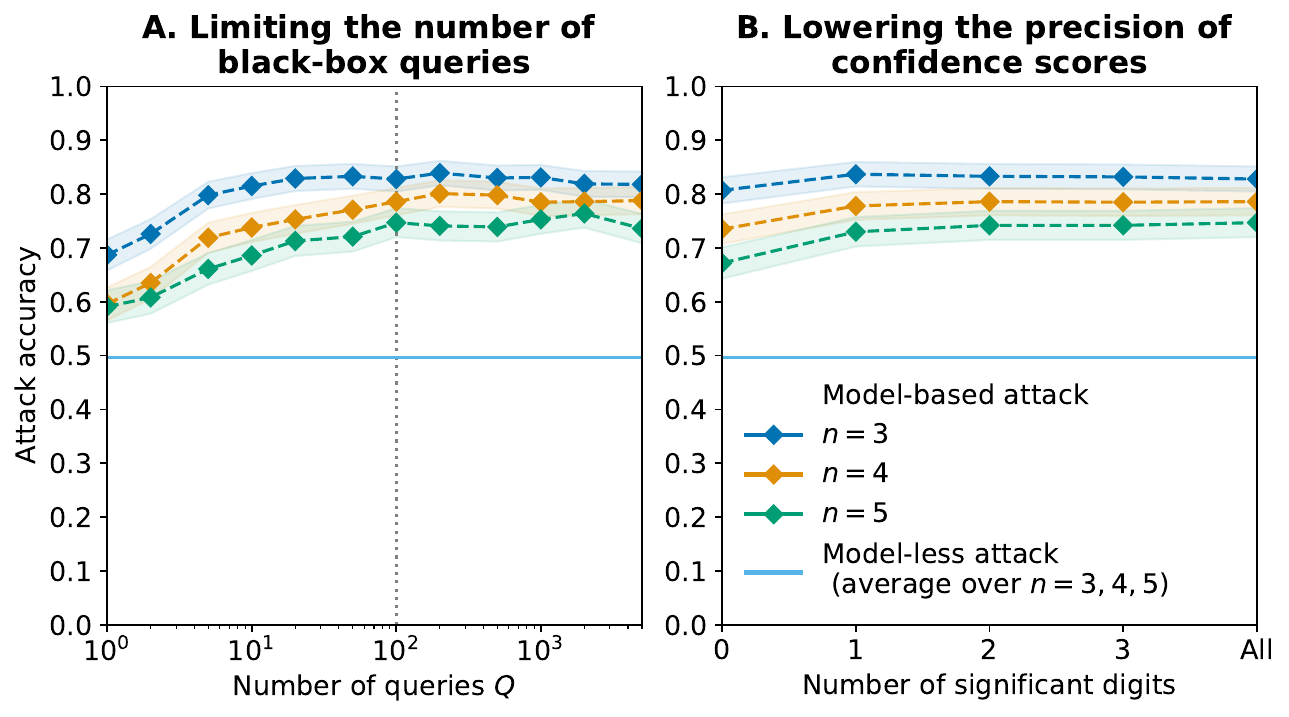}
    \caption{\textbf{Impact of mitigations on the accuracy of our attack against MLP models.} We report results for two mitigations: limiting the number of black-box queries (A) and lowering the precision of confidence scores (B). We report the accuracy (with 95\% confidence interval) of our model-less and model-based attacks over 1000 targets models for $n \in \{3,4,5\}$ variables.}
    \label{fig:mitigations-mlp}
\end{figure}

Fig. \ref{fig:mitigations-mlp} shows the impact of mitigation techniques against our attack targeting MLP models. 

\begin{figure}[!htbp]
\centerline{\includegraphics[width=0.6\textwidth]{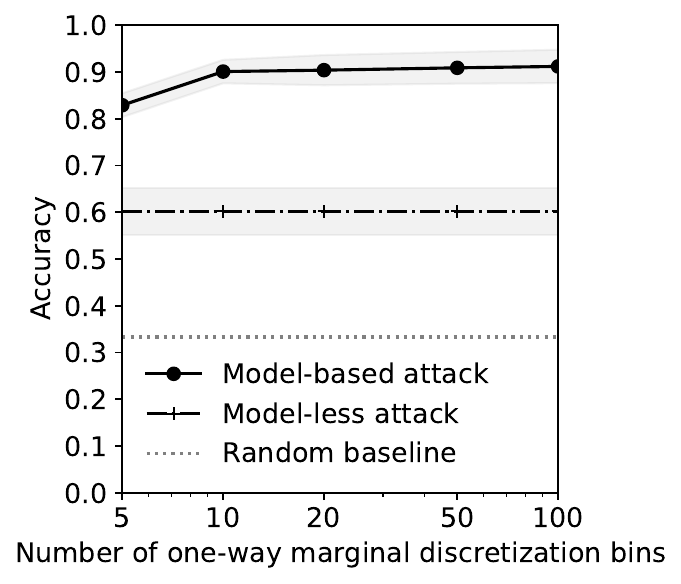}}
\caption{\textbf{Fifa19: Impact of the granularity of one-way marginals available to the attacker.}
We report the attack accuracy (mean and standard deviation) on logistic regression models as we vary the number of discretization bins applied to the one-way marginals.
}
\label{fig:incr_nbr_marginal_bins}
\end{figure}

Fig. \ref{fig:incr_nbr_marginal_bins} shows the impact of the number of sub-intervals $G$ over which the one-way marginals released to the attacker are computed on the accuracy of our attack.

\clearpage
\newpage
\section{Attribute inference attack: details and baselines}
\label{appendix:attribute-inference-attack}
We implement CI-AIA using the same parameters as in the Supplementary Materials~\ref{appendix:experimental-setup:attack-parameters}, $S'=1000$ synthetic datasets, $G=100$ bins, and $m_i=2$ and $\delta_i=0.5$ for every $i=1, \ldots, n-1$.

We now describe the five attacks from the previous works against which we compare our CI-AIA method. 
To perform attribute inference over a continuous sensitive attribute, we divide the attribute range into $G$ bins of equal length, infer the most likely bin using a given attack, then return a uniform sample from the bin as the final prediction. 
Given a candidate record $(x_1^g, x_2, \ldots, x_{n-1}, y)$ where the value of the unknown sensitive attribute $x_1$ is substituted with a sample $x_1^g$ of the $g$-th candidate bin, we will denote by $V_y^g$ the confidence that the candidate record is classified with the target label $y$, i.e., the $l$-th element of output vector $\mathcal{M}_T(x_1^g, x_2, \ldots, x_{n-1})$.

\begin{enumerate}
    \item \textbf{Fredrikson et al.~\cite{fredrikson2014privacy}}
    The attack infers the most likely bin $g=1,\ldots, G$ as per the probability $\text{Pr}(x^g_1) \text{Pr}(\hat{y}=y|\argmax_{l=1}^L \mathcal{M}_T(x_1^g, x_2, \ldots, x_{n-1})=\hat{y})$, where $x^g_1$ is a value sampled uniformly at random in the $g$-th sub-interval.
    The second term is approximated using the confusion matrix of the target model, assumed by Fredrikson et al.~\cite{fredrikson2014privacy} to be known by the attacker.

    \item \textbf{CSMIA (Mehnaz et al.~\cite{mehnaz2022your})} The attack sets the sensitive attribute value of the partial record to each possible value (here, one of $G$ bins) and retrieves the model prediction $\hat{y}^g=\mathcal{M}_T(x_1^g, x_2, \ldots, x_{n-1}), g=1, \ldots, G$.
    Note that we represent a bin using a uniform sample in the bin $x_1^g$.
    Then, the attack distinguishes three cases: (1) if $\hat{y}^g=y$ for only one value of $g$, the approach returns $x_1^g$, (2) if $\hat{y}^g=y$ for more than one value of $g$, the approach returns $x_1^g$ on which the model is the most confident, and (3) if $\hat{y}^g \neq y$ for all values of $g$, the approach returns $x_1^g$ on which the model is the least confident.

    \item \textbf{Yeom et al.~\cite{yeom2018privacy}} The attack returns the sensitive attribute value with the largest prior $\text{Pr}(x_1^g)$ that passes a membership inference test. More specifically, the attack uses a membership oracle $\mathbb{O}$ that returns 1 if $(x_1^g, x_2, \ldots, x_{n-1}, y) \in D_T$ and 0 otherwise.
    The attack returns $\text{argmax}_{g}\text{Pr}(x_1^g)\cdot \mathbb{O}(\mathcal{M}_T, (x_1^g, x_2, \ldots, x_{n-1}))$ (where $x_1^g$ is sampled uniformly at random in the $g$-th bin).
    We use the threshold-based membership oracle of Yeom et al.~\cite{yeom2018privacy}, that returns 1 if and only if the model's confidence on the target record is larger than a threshold $\tau$.
    We use $\tau=0.5$.

    \item \textbf{Jayaraman and Evans~\cite{jayaraman2022attribute}} removes the need for an oracle in the attack by Yeom et al.~\cite{yeom2018privacy}, proposing two variants. The first one is CAI, which returns $\argmax_g V^g_y$. 
    The second one is WCAI, which returns $\argmax_g \text{Pr}(x_1^g | x_2, \ldots, x_{n-1}) V^g_y$, making use of the conditional probability of the sensitive attribute given the non-sensitive attributes. In their work, this quantity is estimated empirically by fitting a machine learning model to infer $X_1$ based on $X_2,\ldots, X_{n-1}$. 
    The model is fitted on auxiliary data drawn from a similar distribution as the private dataset. As we here study a weaker adversary lacking access to auxiliary data, under our assumptions the attack would simply become $\argmax_g \text{Pr}(x_1^g) V_y^g$.

\end{enumerate}

Our correlation inference-based attribute inference attack (CI-AIA) is fundamentally different from Jayaraman and Evans's AIA~\cite{jayaraman2022attribute}.
This is because we do not assume the adversary to know the data distribution via access to auxiliary data.
Instead, we extract information about the distribution from the model (the correlation between the input variables) and use it to generate synthetic data.
In spite of this fundamental difference, the WCAI formula for inferring the sensitive attribute given the partial record is similar to step 3 of our CI-AIA attack.
Recall that in our step 3, we retrieve the average sensitive attribute value $x_1^g$ among all synthetic records which approximately match the partial record $(x_2,x_3,...,x_{n-1})$ and the label $y$. Our step 3 can thus be seen as equivalent to estimating $E(x_1|x_2,...,x_{n-1},y)$, while WCAI returns $\argmax_g p(x_1^g|x_2,...,x_{n-1})V_y^g$.

We study if our CI-AIA can be further improved by applying the WCAI formula instead of our step 3 (described in Sec.~\ref{subsec:materials-methods:attribute-inference}) to the synthetic data generated conditionally on the correlations inferred using our attack.
For completeness, we explore different variants the WCAI formula.

\begin{enumerate}
    \item Variant 1: We trained a regression model for inferring $x_1$ given $x_2,\ldots,x_{n-1}$, like ~\cite{jayaraman2022attribute}. The model is trained on the same synthetic data -- generated conditionally on the correlations inferred from the model -- as our attack, instead of auxiliary data~\cite{jayaraman2022attribute} which our attacker does not have. 
    \item Variant 2: This is the same as variant 1, except that to understand the impact of conditioning on $y$ -- which our step 3 does while~\cite{jayaraman2022attribute} does not -- we trained a regression model for inferring $x_1$ given $x_2, x_3,\ldots,x_{n-1}$, \textbf{and $y$}. 
    \item Variant 3: This is the same as variant 1, except that we multiply $p(x_1^g|x_2,\ldots,x_{n-1})$ by the model's confidence on label $y$, $V_y^g$, to understand if $V_y^g$ helps. This is the closest to the formula of ~\cite{jayaraman2022attribute}, the only difference being how $p(x_1^g|x_2,\cdots,x_{n-1})$ is estimated.
    Jayaraman and Evans~\cite{jayaraman2022attribute} uses a neural network classifier which outputs probabilities, while we use a regression since our attribute $x_1$ is continuous. As the regression model does not output a probability, we model it as the probability of error using a standard normal distribution $p(x_1^g|x_2,\ldots,x_{n-1})\sim e^{-\frac{(x_1^*-x_1^g)^2}{2}}$, where $x_1^*$ denotes the model’s prediction on the partial record $(x_2,\ldots,x_{n-1})$.
    \item Variant 4: This is the same as variant 3, except that to understand the impact of conditioning on $y$ -- which we do while \cite{jayaraman2022attribute} does not -- we trained a regression model for inferring $x_1$ given $x_2, x_3,\ldots,x_{n-1}$ \textbf{and $y$}.
\end{enumerate}

Table \ref{tab:table_aia_variations} shows that none of the variants of our CI-AIA attack achieve better performance than our original attack.
In particular, our CI-AIA achieves the same accuracy (49.7\%) as the WCAI formula~\cite{jayaraman2022attribute} applied to the synthetic data generated using steps 1 and 2 of our attack (49.8\% - Variant 3). 
This means that once correlations have been extracted and synthetic data has been generated conditionally on these correlations using steps 1 and 2 of our attack, either our step 3 or Jayaraman and Evans's~\cite{jayaraman2022attribute} formula can be used to achieve similar results. 
Another interesting finding is that using $y$ in the prediction leads to better performance than not using $y$ at all.
The AIA accuracy is 49.5\% when $x_1$ is inferred conditionally on $y$ in addition to the other variables (Variant 2), 49.8\% when the prediction uses the model's confidence on label $y$ (Variant 3) and 46.7\% when $y$ is not used at all (Variant 1). 
Using $y$ by both conditioning on it and via the model's confidence (Variant 4) does not significantly improve AIA performance.

\begin{table}[!htbp]
    \centering
    \begin{tabular}{lc}
    \toprule
         \textbf{Method}& \textbf{Accuracy} \\
    \midrule
    (Ours) CI-AIA & 49.7 $\pm$ 1.0  \\
    \midrule
    Variant 1 & 46.7 $\pm$ 1.0\\
    Variant 2 & 49.5 $\pm$ 1.0\\
    Variant 3 (closest formula to \cite{jayaraman2022attribute}) & 49.8 $\pm$ 1.0\\
    Variant 4 & 50.5 $\pm$ 0.9\\
    \bottomrule
    \end{tabular}
    \caption{\textbf{Comparison between our CI-AIA and attacks obtained by substituting Jayaraman and Evans's~\cite{jayaraman2022attribute} prediction rule to step 3 of CI-AIA.} We report the attack accuracy  averaged over 1000 runs (with 95\% confidence interval) on the Fifa19 dataset.}
    \label{tab:table_aia_variations}
\end{table}

\end{document}